\documentclass{article}

\usepackage{microtype}
\usepackage{graphicx}
\usepackage{subfig}
\usepackage{booktabs} 
\usepackage{amsmath,amssymb, amsfonts, mathtools}
\usepackage{amsthm}
\usepackage{hyperref}
\usepackage{xspace}

\usepackage{cleveref}
\crefname{algocf}{alg.}{algs.}
\Crefname{algocf}{Algorithm}{Algorithms}
\crefname{section}{\S}{\S}

\usepackage{multicol}
\usepackage{multirow}
\usepackage{stfloats}
\usepackage{float}
\usepackage{enumitem}
\usepackage{tikz}

\usepackage{setspace}
\usepackage{textcomp}
\usepackage{xcolor}
\usepackage{subfig}
\usepackage{url}
\usepackage{array}
\usepackage{makecell}
\usepackage{bbm}

\usepackage{bm}
\usepackage[algo2e,linesnumbered,lined,ruled,commentsnumbered]{algorithm2e}

\usepackage{ifthen}
\usepackage[normalem]{ulem} 
\usepackage{xcolor}
\usepackage{amssymb}

\newboolean{showedits}
\setboolean{showedits}{true} 
\ifthenelse{\boolean{showedits}}
{
	\newcommand{\del}[1]{\textcolor{red}{\sout{#1}}} 
}{
	\newcommand{\del}[1]{} 
	
}

\newboolean{showcomments}
\setboolean{showcomments}{true} 
\newcommand{\id}[1]{$-$Id: scgPaper.tex 32478 2010-04-29 09:11:32Z oscar $-$}

\ifthenelse{\boolean{showcomments}}
{\newcommand{\nbc}[3]{
 {\colorbox{#3}{\bfseries\sffamily\scriptsize\textcolor{white}{#1}}}
 {\textcolor{#3}{\sf\small$\blacktriangleright$\textit{#2}$\blacktriangleleft$}}}
 }
{\newcommand{\nbc}[3]{}
 \renewcommand{\del}[1]{} 
 }

\definecolor{ibcolor}{rgb}{0.9,0.5,0}
\definecolor{dsrcolor}{rgb}{0.5,0.6,0}
\definecolor{cfcolor}{rgb}{0,0.5,0.9}
\definecolor{lwcolor}{rgb}{0.2,0.8,0.4}
\definecolor{eycolor}{rgb}{0.7,0.6,1.0}
\definecolor{oldcolor}{rgb}{0.2,0.2,0.2}
\definecolor{tdcolor}{rgb}{0.0,0.5,0.7}

\newcommand\shiqi[1]{\nbc{SH}{#1}{dsrcolor}}
\newcommand\mathias[1]{\nbc{ML}{#1}{cfcolor}}

\newcommand{\mytitle}{GlueFL\xspace}
\newtheorem{assumption}{Assumption}
\newtheorem{theorem}{Theorem}
\newtheorem{proposition}{Proposition}
\newtheorem{lemma}{Lemma}

\crefname{assumption}{assumption}{assumptions}

\usepackage[accepted]{mlsys2023}

\mlsystitlerunning{
\mytitle: Reconciling Client Sampling and Model Masking for Bandwidth Efficient Federated Learning
}

\begin{document}

\twocolumn[
\mlsystitle{
\mytitle: Reconciling Client Sampling and Model Masking for Bandwidth Efficient Federated Learning
}



\mlsyssetsymbol{equal}{*}

\begin{mlsysauthorlist}
\mlsysauthor{Shiqi He}{ubc}
\mlsysauthor{Qifan Yan}{ubc}
\mlsysauthor{Feijie Wu}{purdue}
\mlsysauthor{Lanjun Wang}{tju}
\mlsysauthor{Mathias Lécuyer}{ubc}
\mlsysauthor{Ivan Beschastnikh}{ubc}
\end{mlsysauthorlist}

\mlsysaffiliation{ubc}{University of British Columbia}
\mlsysaffiliation{purdue}{Purdue University}
\mlsysaffiliation{tju}{Tianjin University}

\mlsyscorrespondingauthor{Shiqi He}{shiqihe@cs.ubc.ca}
\mlsyscorrespondingauthor{Ivan Beschastnikh}{bestchai@cs.ubc.ca}

\mlsyskeywords{Machine Learning, MLSys}

\vskip 0.3in

\begin{abstract}
Federated learning (FL) is an effective technique to directly involve edge devices in machine learning training while preserving client privacy. However, the substantial communication overhead of FL makes training challenging when edge devices have limited network bandwidth. Existing work to optimize FL bandwidth overlooks downstream transmission and does not account for FL client sampling. 

In this paper we propose \emph{\mytitle}, a framework that incorporates new client sampling and model compression algorithms to mitigate low download bandwidths of FL clients. \mytitle prioritizes recently used clients and bounds the number of changed positions in compression masks in each round. Across three popular FL datasets and three state-of-the-art strategies, \mytitle reduces downstream client bandwidth by 27\% on average and reduces training time by 29\% on average.

\end{abstract}
]



\printAffiliationsAndNotice{}  
\allowdisplaybreaks

\section{Introduction}
\label{sec:introduction}
Federated learning (FL) moves machine learning (ML) training to the edge. In FL, edge clients communicate with a central server to collaboratively train a global model, while keeping client training data local.
We focus on \emph{cross-device} FL, in which there are many clients that are end-user devices. For example, companies like Google and Intel use cross-device FL for computer vision and natural language processing model training across customer devices~\cite{hartmann2019federated, yang2018applied, hard2018federated}.



One downside of FL is its network usage. This is especially problematic in cross-device FL, which relies on lower-bandwidth mobile or IoT devices~\cite{kairouz2021advances}. 
For example, Google Keyboard (Gboard), a virtual keyboard with over 1 billion installs, selects clients from millions of mobile devices to enhance its search query suggestions~\cite{yang2018applied}. In this type of application, clients usually have a diversity of device-to-server (upstream) and server-to-device (downstream) bandwidth. Clients that have either slow upstream or downstream bandwidth act as stragglers and slow down model training. 
%

This heterogeneous bandwidth setting has attracted significant research, with a focus on reducing the communication cost of FL training~\cite{chen2021communication,sattler2019robust,vargaftik2022eden, reisizadeh2020fedpaq, mcmahan2017communication}. One important strategy is \emph{client sampling}, which limits the number of clients that perform training in each round~\cite{mcmahan2017communication, luo2022tackling}. 
Client sampling reduces both upstream and downstream bandwidth. However, a client that is not sampled gradually becomes stale: its local state diverges from the state of clients that have been sampled.
The next time this client is sampled, the central server must therefore send a larger state update, increasing the downstream transmission overhead.
%
%

Another approach to reducing FL bandwidth usage is to apply a mask to the client gradients, such as a \textit{sparsification} mask~\cite{sattler2019robust, wangni2018gradient} or a \textit{parameter freezing} mask~\cite{chen2021communication,brock2017freezeout}. In traditional masking schemes, clients apply a mask to their local gradients and only transfer significant gradients to the server. This saves upstream bandwidth. Since each client generates the mask locally and independently, however, the entire model is usually updated at the end of a round and needs to be fully synchronized. 
In \emph{server masking} schemes, such as Sparse Ternary Compression (STC)~\cite{sattler2019robust} and Adaptive Parameter Freezing (APF)~\cite{chen2021communication},
%
%
the server uses a mask to compute the final model update. Since the server only partially updates the model, only a part of the model needs to be sent back to clients; this saves downstream bandwidth.

User sampling and masking approaches are typically considered as orthogonal, compatible approaches~\cite{sattler2019robust,chen2021communication}.
Though existing masking strategies are indeed empirically effective in full participation FL, we show that when client sampling is used they fail to decrease \emph{downstream} bandwidth (\S\ref{sec:background}). For example, with a 0.01 sample ratio and a masking compression ratio of 10\%, a single client needs to download 75\% of the global model on average. 
%
%
%
The reason downstream bandwidth increases is because of \emph{the staleness of local state at the clients}.
To see why, let us first consider the full participation case.
Intuitively, since the global model is only partially updated by the server under masking, a client only needs to download this partial update and apply it to its local version of the model, saved from the previous round. 
With client sampling however, a typical client skips multiple rounds by not being sampled, and its local model state becomes stale. When the client is later sampled, it needs to download the new value of all parameters updated in the skipped rounds, which amounts to a large fraction of the model. This effect increases downstream bandwidth usage, voiding the benefits of server masking, and slowing down training when edge devices have limited download capacity~\cite{speedtest}. 

To resolve the incompatibility between masking and client sampling, we propose \textbf{\mytitle}, a new FL training framework specifically designed to retain the benefits of masking when using client sampling.
%
%
This compatibility is particularly important in cross-device FL deployments, which require both client sampling (full participation is impractical) and bandwidth savings due to mobile or IoT clients.
%
To the best of our knowledge, \mytitle is the first masking design to address the downstream bandwidth bottleneck in cross-device FL with client sampling.

%
%
%

We design \mytitle with two new mechanisms to alleviate client staleness and to optimize downstream bandwidth requirements. First, we introduce \emph{sticky sampling} (\S\ref{sec:sticky_sampling}) to prioritize the most recently used clients, thereby reducing the number of stale clients in each update.
Since recently selected clients have an up-to-date view of model parameters, they need to download smaller updates.
We combine sticky sampling with a weighted central aggregation scheme to ensure that model updates remain unbiased, a requirement for convergence (\S\ref{sec:convergence}).
Sticky sampling is especially important in practical implementations that sample a small fraction of clients in each round~\cite{yang2018applied}.

Second, we propose a gradual \emph{mask shifting} strategy (\S\ref{sec:mask-shiftnig}), to ensure that consecutive central model updates share a large number of changed parameters, while empirically preserving model convergence. 
This way, a newly selected client only has to synchronize a subset of the model, even after several rounds of not being sampled.

To sum up, we make three contributions:
\begin{itemize}[leftmargin=*,topsep=0pt]  \renewcommand{\labelitemi}{$\star$}
    \itemsep 0em
    \item We present an FL design called \mytitle,  which is based on sticky sampling and mask shifting. These two new mechanisms alleviate the impact of client staleness in client sampling. Both techniques minimize downstream bandwidth in cross-device FL. To the best of our knowledge, this is the first work to combine masking with client sampling to reduce downstream bandwidth.
    %
    %
    \item We analyse FL convergence under \mytitle's sticky sampling, and show that our proposed weighted aggregation preserves unbiasedness of updates and convergence. 
    \item We evaluate \mytitle empirically, and demonstrate downstream bandwidth and training time savings on three public datasets. On average, our evaluation shows that \mytitle spends 29\% less training time with a 27\% less downstream bandwidth overhead as compared to FedAvg~\cite{mcmahan2017communication}, STC~\cite{sattler2019robust} and APF~\cite{chen2021communication}.
    %
\end{itemize}

\begin{table}[t] 
\footnotesize
\centering
\setstretch{1}
\caption{Summary of notation used in this paper.
}
\begin{tabular}{cl}
\hline\rule{0pt}{2.2ex} 
$\mathcal{N}, N, i$ & set, total number, index of \emph{clients} \\
$\mathcal{K}, K$ & set, number of \emph{sampled clients}\\
$T, t$ & number, index of \emph{communication rounds} \\
$E, e$ & number, index of \emph{local update steps} \\ 
$\mathbf{w}^t$ & server model in round $t$ \\
$\mathbf{w}^{t,e}_i, \mathbf{g}^{t,e}_i$ & model, gradients of client $i$ in round $t$ and step $e$\\
$\mathcal{S}, S$ & set, size of \emph{sticky group} \\
$\mathcal{C}, C$ & set, number of clients sampled from $\mathcal{S}$\\
$\mathcal{R}, R$ & set, number of clients sampled from $\mathcal{N} \setminus \mathcal{S}$\\
$\nu_{i,s}, \nu_{i,r}$ & aggregation weight of client $i$ in $\mathcal{C}$, $\mathcal{R}$\\
$q, q_{shr}$ & total, shared \emph{mask ratio} \\
\hline
\end{tabular}
\label{table:notation}
\end{table}

\section{Motivation and Background} 

\label{sec:background}
We start by reviewing standard FL with client sampling. Then we introduce a state of the art masking strategy called STC~\cite{sattler2019robust}, and discuss its limitations. Finally, we formalize the problem that we set out to solve in the rest of the paper. \Cref{table:notation} overviews our notation.

\subsection{Federated Learning (FL)}
\label{sec:background_fl}
Consider a system with $N$ clients, coordinated by a central server. Each client $i$ has a local data distribution $\mathcal{D}_i$. Let us denote the weight of client $i$ as $p_i$ such that $\sum_{i=1}^{N} p_i = 1$. The weight $p_i$ is given by the server and represents the importance of the $i$-th client's local loss function. Under the non-convex settings, our target is formulated as 
\begin{equation} \label{equation:problem}
    \min_{\mathbf{w} \in \mathbb{R}^d} \quad F(\mathbf{w}) \overset{\triangle}{=} \sum_{i=1}^{N} p_i F_i(\mathbf{w})
\end{equation}
where $F_i(\mathbf{w}) = \frac{1}{|\mathcal{D}_i|} \sum_{\xi \in \mathcal{D}_i} \ell(\mathbf{w}, \xi)$, and $\ell(\mathbf{w}, \xi)$ is the empirical loss on model $\mathbf{w}$ and sample $\xi$. In practice, $F_i(\mathbf{w})$ is generally estimated with a random realization $\xi_i$ drawn from $\mathcal{D}_i$, which is assumed to be unbiased, i.e., $\mathbb{E}_{\xi_i \sim \mathcal{D}_i} \ell(\mathbf{w}, \xi_i) = F_i(\mathbf{w})$. Let $F_*$ is the minimum value of the global objective, i.e., $F(\mathbf{w}) \geq F_*$ for any $\mathbf{w} \in \mathbb{R}^d$. 

FedAvg~\cite{mcmahan2017communication} is a standard algorithm to solve \Cref{equation:problem}. To improve communication efficiency, clients are selected uniformly at random in each round. The FedAvg algorithm with client sampling looks as follows:
\begin{enumerate}
    \itemsep 0em
    \item At the beginning of round $t$, the server uniformly at random samples a subset of clients (i.e., $\mathcal{K}$) and broadcasts the latest global model $\mathbf{w}^t$ to these sampled clients.
    \item Each sampled client $i \in \mathcal{K}$ receives the model $\mathbf{w}^t$ ($=\mathbf{w}^{t,0}_i$) and runs $E$ local SGD iterations to compute a local update $\Delta^{t}_i=-\gamma\sum_{e=0}^{E-1} g^{t,e}_i$, where $\gamma$ is the client learning rate. In each iteration, the client computes the gradient as $g^{t,e}_i=\nabla \ell(\mathbf{w}_i^{t, e}, \xi_i^{t, e})$ where $\xi_i^{t, e}$ is drawn from $\mathcal{D}_i$.
    \item The server receives updates $\Delta^{t}_i$ from all sampled clients and aggregates them to compute the new global model \cite{li2019convergence}
    \begin{equation} \label{eq:fedavg_update}
        \mathbf{w}^{t+1} = \mathbf{w}^{t} + \frac{N}{K} \sum_{i \in \mathcal{K}} p_i \Delta^{t}_i    
\end{equation}
\end{enumerate}
In expectation, the steps above realize an update form $\mathbb{E}_{\mathcal{K}} \big[\mathbf{w}^{t+1}\big] = \mathbf{w}^t + \sum_{i=1}^N p_i \Delta_i^{t}$ in each round. 
To ensure that the global loss approaches the optimal one, FedAvg repeats the process for $T$ rounds. FedAvg achieves a convergence rate of $O\left(\sqrt{\frac{E}{KT}}\right)$ \cite{karimireddy2020scaffold, yang2021achieving} under partial worker participation. 

\setlength{\textfloatsep}{0.1cm}
\begin{algorithm2e}[t]
\footnotesize
\setstretch{0.8}
\SetKwInOut{output}{Output}
\output{$\mathbf{w}^T$}
\For{$t \leftarrow 1$ \KwTo $T$}{
    \CommentSty{/* Server:client sampling */} \\
    Generate set of sampled clients $\mathcal{K}$ \; 
    Broadcast $\mathbf{w}^t$ to $\mathcal{K}$ \;
    \CommentSty{/* Client:local training */} \\
    \For{$i \in \mathcal{K}$ in parallel }{
        $\mathbf{w}^{t,0}_i \gets \mathbf{w}^{t}$ \;
        \For{$e \leftarrow 0$ \KwTo $E - 1$}{
            $\mathbf{w}^{t,e+1}_i \gets \mathbf{w}^{t,e}_i - \gamma \mathbf{g}^{t,e}_i$;
        }
        \CommentSty{/* Client:sparsification */} \\
        $\Tilde{\Delta}_i^{t} \gets top_{q }(\mathbf{w}^{t,E}_i - \mathbf{w}^{t,0}_i)$ \; \label{stc-line:sparsification}
    }
    \CommentSty{/* Server:aggregation */} \\
    Receive $\Tilde{\Delta}_i^{t}$ from worker $i \in \mathcal{K}$ \;
    \CommentSty{/* Server:sparsification */} \\
    $\Tilde{\Delta}^{t} \gets top_{q }(\sum_{i \in \mathcal{K}} p_i \frac{N}{K} \Tilde{\Delta}_i^{t})$  \; \label{stc-line:server-sparsification}
    $\mathbf{w}^{t+1} \gets \mathbf{w}^{t} + \Tilde{\Delta}^{t}$;
}
\caption{Sparse Ternary Compression (STC)}
\label{algo:stc}
\end{algorithm2e}
\normalsize

\subsection{Cross-device FL bandwidth characteristics}
\label{sec:bandwidth}

\begin{figure}[!t]
\centering
\vspace{-15px}
\subfloat[]{%
\centering
\includegraphics[width=0.48\linewidth]{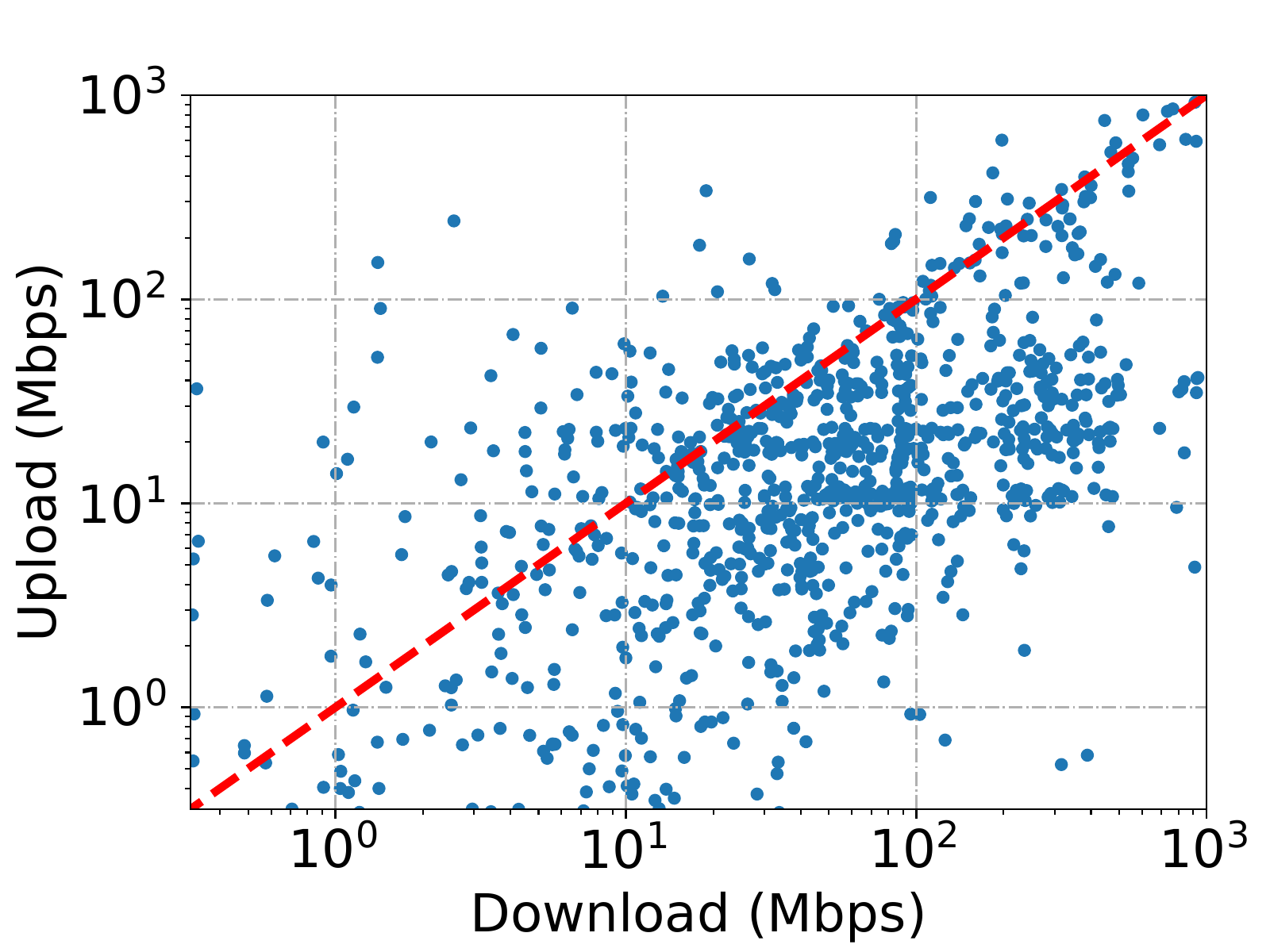}
\label{fig:bandwidth_dist}
}%
\subfloat[]{%
\centering
\includegraphics[width=0.48\linewidth]{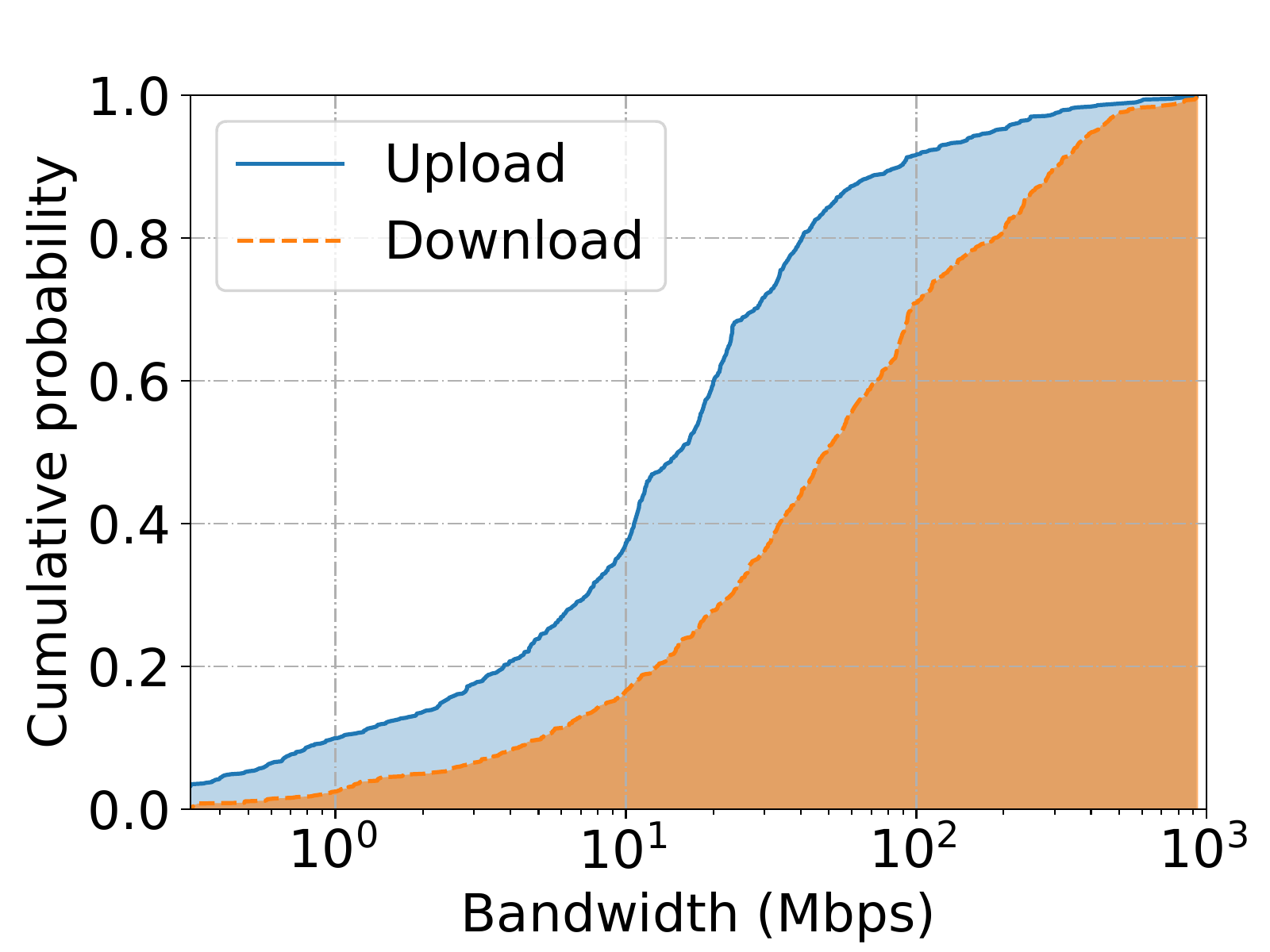}%
\label{fig:bandwidth_cdf}
}%
\caption{(a) The distribution of network bandwidth in North America, June 2022~\cite{mlab}, and (b) the cumulative distribution function (CDF) of network bandwidth in (a).}

\label{fig:bandwidth_na}
\centering
\vspace{-30px}

\end{figure}

The cross-device FL setting relies on a large number of clients. In this case, some clients are likely to have an unreliable or slow network. For example, \Cref{fig:bandwidth_na} shows the bandwidth distribution estimated by measurement lab~\cite{mlab}. We observe that around 20\% of devices have a download bandwidth of at most 10Mbps. These devices can take at least 20s to download a typical model like ShuffleNet\_V2~\cite{zhang2018shufflenet}, which is specially designed for mobile devices and contains 5 million model parameters. 
%

\subsection{Limitations of Existing Masking Strategies}
\label{sec:limitations}

\begin{figure}[!t]
\subfloat[]{%
\centering
\includegraphics[width=0.48\linewidth]{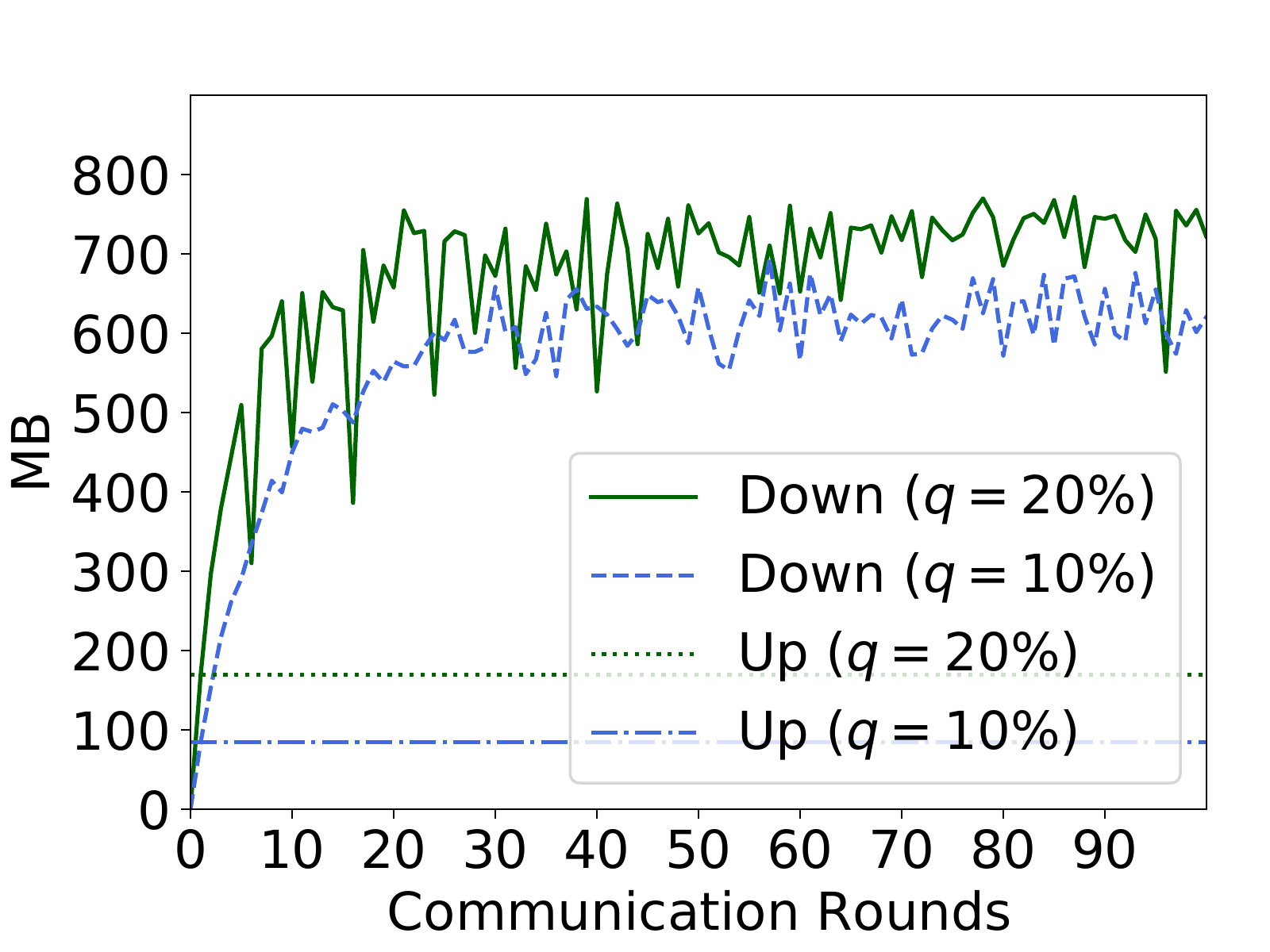}
\label{fig:network_bandwidth}
}%
\subfloat[]{%
\centering
\includegraphics[width=0.48\linewidth]{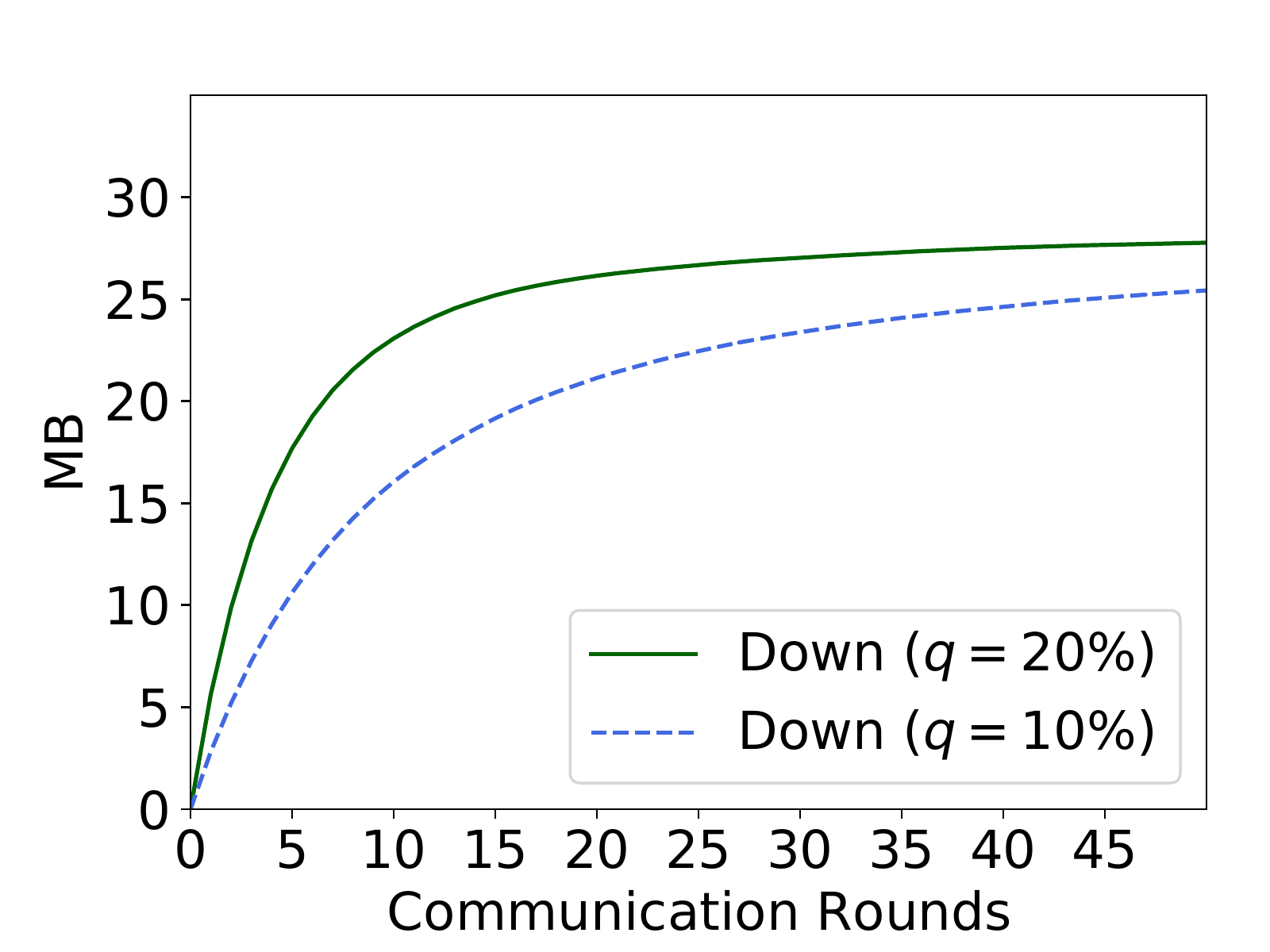}%
\label{fig:resnet_bandwidth}
}%
\centering
\caption{(a) The downstream and upstream bandwidth usage of STC per round, and (b) the model size a client must download when being re-sampled after a certain number of rounds.} 
\label{fig:exp_prelim}
\end{figure}


Prior work has proposed several masking strategies to reduce the amount of transferred data and alleviate low bandwidth issues~\cite{sattler2019robust, wangni2018gradient,chen2021communication,brock2017freezeout}.
To demonstrate how masking fails to optimize downstream bandwidth in FL with client sampling, we use STC~\cite{sattler2019robust}, a popular server masking strategy, as a representative technique. 

STC builds on~\citet{stich2018sparsified}, a masking approach that selects and uploads the largest $q$ (e.g., $10\%$) absolute values in a client's local gradients. In STC, this top-k sparsification technique is applied to both clients' gradients and server updates\footnote{
For simplicity, we only consider the masking part of STC---STC also includes quantization, an orthogonal technique that can be combined with sparsification~\cite{jiang2018linear, basu2019qsparse} and will not change our conclusion, as quantization compresses both downstream and upstream communication.}.
\Cref{algo:stc} 
shows this masking-only version of STC. For a single client sampled in both the current and last round, STC only has to update the weights covered by the server mask (line~\ref{stc-line:server-sparsification}). However, note that a client that has \emph{not} been sampled recently may have to update the entire model, as their local view of the model is stale. The reason is that server masks change in each round, and the client has to synchronize all updated model parameters since it last participated.

To measure the impact of model staleness on downstream bandwidth, we apply STC to FedAvg and conduct experiments on FEMNIST, using $N=2,800$ clients and a client sample size of $K=30$. We try compression ratios of 10\% and 20\%\footnote{Smaller values led STC to require an unacceptable number of rounds to converge with a noticeable drop in convergence accuracy.}.
We examine both downstream and upstream bandwidth usage in each round. The results in~\Cref{fig:exp_prelim} show that upstream bandwidth is reduced when using a smaller compression ratio, as expected. 
However, a client still needs to download 70\% of the global model on average. Clients with 10Mbps download bandwidth (\S\ref{sec:bandwidth}) will take at least 14s to receive these changes. This imposes a high downstream bandwidth requirements on participating clients. 
In general, the more rounds that a client skips, the more updated model state it needs to download (\Cref{fig:resnet_bandwidth}). As a result, the training bottleneck shifts to downstream communication. 
%
%
We expect these results to hold for other masking strategies as they similarly update different parts of the global model in each communication round. For example, in APF~\cite{chen2021communication}, model parameters are frozen in some rounds but will then be updated again after the freezing period ends. The downstream bottleneck is therefore a general limitation across masking strategies.


\vspace{-7px}
\subsection{Problem Setup}

Our goal in \mytitle is to minimize the total expected downstream bandwidth of training, while retaining a low upstream bandwidth, and 
ensuring that the expected global training loss $F(\mathbf{w}^T)$ converges to a local minimum value, where $\mathbf{w}^T$ is the aggregated global model after $T$ rounds. 


\vspace{-5px}
\section{\mytitle Framework Design} 
\label{sec:framework}
\mytitle includes two components to decrease the downstream bandwidth during FL training: sticky sampling (\Cref{fig:client-sampling}) and mask shifting (\Cref{fig:mask-shifting}). The newly designed sampling scheme allows some clients to be re-sampled in a short term and mask shifting restricts the mask from changing too fast. 
%
We elaborate on the design of each of these components in \S\ref{sec:sticky_sampling} and \S\ref{sec:mask-shiftnig}, before describing how to adapt other existing mechanisms in \S\ref{sec:other-tweaks}.

\vspace{-5px}

\subsection{Sticky Sampling}
\label{sec:sticky_sampling}

Client sampling is the process of selecting $K$ out of $N$ clients in each round, to participate in computing the model update. With uniform sampling, each client participates in each round with a probability of $K/N$. Thus, a client is expected to participate in training every $N/K$ rounds on average (See \Cref{prop:uniform} 
in Appendix~\ref{proof:propsticky}). In cross-device FL systems, the value of $N$ is often large, and $K$ is small. 
For example, Gboard samples $K=100$ clients in each round while there are millions of devices~\cite{yang2018applied}.
%
%
This produces a low probability of participation in each round, which means that on average clients skip a large number of training rounds before being selected again. As we saw in \S\ref{sec:limitations}, these long skips are responsible for local state staleness. Clients' state must therefore be re-synchronized when they are selected, reducing the benefits of masking on downstream bandwidth.
%

\begin{figure}[t]
    \centering
    \includegraphics[width=1.0\linewidth]{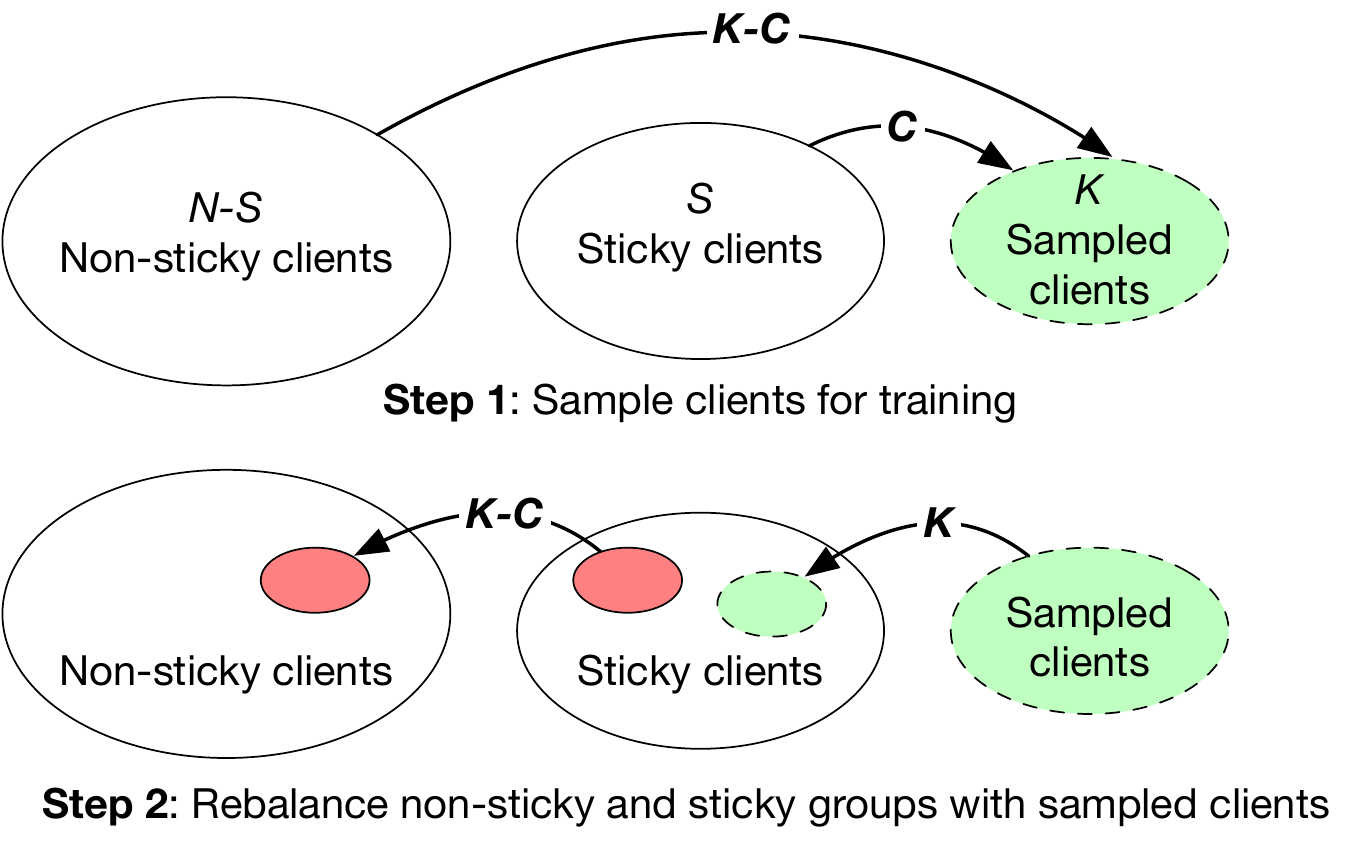}
    \caption{Sticky sampling design. }
    \label{fig:client-sampling}
\end{figure}

\mytitle introduces \emph{sticky sampling} to ensure that clients with an up-to-date local state are more likely to be selected.
%
\Cref{fig:client-sampling} illustrates sticky sampling and \Cref{algo:sticky_sampling} details it. The server maintains a smaller \emph{sticky group} of clients $\mathcal{S}$ with size $S$, while the remaining clients form a \emph{non-sticky} group, $\mathcal{N} \setminus \mathcal{S}$.  We randomly select $S$ clients to initialize $\mathcal{S}$ in the beginning of training, and allow $\mathcal{S}$ to evolve over time.
%

\Cref{fig:client-sampling} (step 1) illustrates how in each FL training round, the server constructs its sampled set of clients $\mathcal{K}$ from two sources; $\mathcal{K}=\mathcal{C} \cup \mathcal{R}$. It samples $C$ clients to construct $\mathcal{C}$ by sampling from the current sticky group $\mathcal{S}$. It samples $(K-C)$ clients to construct $\mathcal{R}$ by sampling from the non-sticky group, without replacement. All sampled clients $\mathcal{K}$ participate in one round of training (\Cref{algo:sticky_sampling} line~\ref{algo:sticky_sampling_sampling}).

At the end of the round (\Cref{fig:client-sampling} (step 2)), the server randomly selects $(K-C)$ clients from $\mathcal{S} \setminus \mathcal{C}$ (the set of clients in the sticky group that did not participate in the latest round) and removes these clients from the sticky group (\Cref{algo:sticky_sampling} line~\ref{algo:sticky_sampling_update}). The server replaces these clients with $(K-C)$ clients that were \emph{not} sampled from the sticky group and that participated in the last update ($\mathcal{R}$ in \Cref{algo:sticky_sampling}).

Just as with uniform client sampling, sticky client sampling requires $N/K$ rounds 
to re-sample a client on average (see \Cref{prop:sticky} for details). However, a client selected with sticky sampling will join the sticky group and then have a higher probability of being selected in the next round than under uniform sampling, as long as $\frac{C}{S} > \frac{K}{N}$.
Since a client that exits the sticky group (by not being selected in the current round) is less likely to be selected than under uniform sampling ($\frac{K-C}{N-S} < \frac{K}{N}$ when $\frac{C}{S} > \frac{K}{N}$), we need to ensure that a sticky client has a higher expectation of being included during the next several rounds. This is because after several missed rounds, the whole model needs to be synchronized (see \Cref{fig:resnet_bandwidth}). \Cref{prop:sticky} in Appendix \ref{proof:propsticky} shows the probability for a client in the sticky group to be selected after $r$ rounds. We use this formula to select $S$ and $C$ to ensure that this probability is higher than that of uniform sampling for a high enough value of $r$.

\textbf{Case Study.} Consider a training run on FEMNIST with $N=2,800$ clients, $K=30$, $S=120$, and $C=24$ (our default experimental setup in \S\ref{sec:experimental_setup}). 
In this case, using the \Cref{prop:uniform} and \Cref{prop:sticky} in Appendix, we can compute the probability of client inclusion over the next 6 rounds for a client starting in the sticky group:
$20.0\%, 15.0\%, 11.2\%, 8.5\%, 6.4\%, 4.8\%$. By contrast, uniform sampling re-samples clients with a probability of around $1.1\%$.

With sticky sampling, clients that just participated in a round, and thus have an up-to-date state, are more likely to participate again in the short term. Such clients will therefore download smaller model updates.
This synergizes with masking approaches that reduce the size of an update in each round.
We show in \S\ref{sec:experiment} that for cross-device FL, where a large $N$ and a small $K$ are typical, masking approaches with sticky sampling significantly reduce downstream bandwidth usage.


However, sticky sampling also introduces new challenges during aggregation.
As discussed in \S\ref{sec:background}, the global update should provide appropriate representation for every client in expectation~\cite{wang2020tackling, mitra2021linear,wu2021deterioration}. Formally, the update should be an unbiased estimate of the FedAvg update computed on every client in round $t$. That is: $\mathbb{E}_\mathcal{K} [\Delta^t] = \sum_{i=1}^N p_i \Delta_i^t$.
%
Under the FedAvg aggregation function (\Cref{eq:fedavg_update}),
since sticky clients are selected with higher probability, they would have a larger weight then non-sticky clients. To correct for this bias, \mytitle uses an inverse propensity weighted aggregation function. It assigns a different weight to updates from clients of different groups, corresponding to their importance parameter re-weighted by the inverse probability of selection. Updates from sticky group clients use the weight $\nu_{i, s}^t=\frac{S}{C}p_i$, while non-sticky group clients use the weight $\nu_{i, r}^t=\frac{N-S}{K-C}p_i$. 
The model update rule then becomes: 
\begin{equation}
    \label{eq:sticky_global_update}
    \mathbf{w}^{t+1} \gets \mathbf{w}^{t} + \underbrace{\sum_{i \in \mathcal{C}} \nu_{i, s}^t \cdot \Delta^{t}_i + \sum_{i \in \mathcal{R}} \nu_{i, r}^t \cdot \Delta^{t}_i}_{\Delta^t}
\end{equation}
This is shown in lines~\ref{algo:sticky_sampling_aggregation_1} and~\ref{algo:sticky_sampling_aggregation} of Algorithm~\ref{algo:sticky_sampling}. 


\setlength{\textfloatsep}{0.1cm}
\begin{algorithm2e}[t]
\footnotesize
\SetKwInOut{output}{Output}
\output{$\mathbf{w}^T$}
\For{$t \leftarrow 1$ \KwTo $T$}{
\CommentSty{/* Server:sample clients */} \\
Randomly select $|\mathcal{C}| = C$ clients from $\mathcal{S}$\; 
Randomly select $|\mathcal{R}| = K - C$ clients from $\mathcal{N} \setminus \mathcal{S}$\; 
Set of sampled clients $\mathcal{K} \gets \mathcal{C} \cup \mathcal{R}$\; \label{algo:sticky_sampling_sampling}
Broadcast $\mathbf{w}^t$ to $\mathcal{K}$\;
\CommentSty{/* Client:local training */}\\
\For{$i \in \mathcal{K}$ in parallel }{
    $\mathbf{w}^{t,0}_i \gets \mathbf{w}^{t}$ \;
    \For{$e \leftarrow 0$ \KwTo $E - 1$}{
        $\mathbf{w}^{t,e+1}_i \gets \mathbf{w}^{t,e}_i - \gamma \mathbf{g}^{t,e}_i$;
    }
    $\Delta_i^{t} \gets \mathbf{w}^{t,E}_i - \mathbf{w}^{t,0}_i$ \;
} 
\CommentSty{/* Server:aggregation */} \\
Receive $\Delta_i^{t}$ from worker $i \in \mathcal{K}$\;
$\Delta^{t} \gets \sum_{i \in \mathcal{C}} \nu_{i, s}^t \Delta^{t}_i + \sum_{i \in \mathcal{R}} \nu_{i, r}^t \Delta^{t}_i$ \; \label{algo:sticky_sampling_aggregation_1}
$\mathbf{w}^{t+1} \gets \mathbf{w}^{t} + \Delta^{t}$\; \label{algo:sticky_sampling_aggregation}
\CommentSty{/* Server:rebalance non-sticky and sticky groups */} \\ 
Randomly remove $K - C$ clients in $\mathcal{S} \setminus \mathcal{C}$ \; \label{algo:sticky_sampling_update}
$\mathcal{S} \gets \mathcal{S} \cup \mathcal{R}$ \;
}
\caption{Sticky Sampling}
\label{algo:sticky_sampling}
\end{algorithm2e}
\setlength{\floatsep}{0.1cm}
\normalsize

With this reweighting scheme in place, we can show that sticky sampling udpates are unbiased:

\newcommand{\w}{\mathbf{w}}
\newcommand{\g}{\mathbf{g}}
\newcommand{\x}{\mathbf{x}}
\newcommand{\m}{M}
\newcommand{\E}{\mathbb{E}}
\newcommand{\1}{\mathbbm{1}}
\newcommand{\norm}[1]{\left\|#1\right\|_2^2}
\newcommand{\bracket}[1]{\left(#1\right)}
\newcommand{\conE}{\mathbb{E}_{t+1 | t}}
\newcommand{\q}{\mathcal{Q}}

\begin{theorem}[Unbiased Aggregation]
\label{theorem:unbiased}
Let $\mathcal{K}=\mathcal{C} \cup \mathcal{R}$ be the set of sampled clients in sticky sampling. The update $\Delta^t$ computed in \Cref{eq:sticky_global_update} is unbiased. That is:
\begin{equation}
    \E_{\mathcal{K}} [\Delta^{t}] = \sum_{i=1}^N p_i \Delta^{t}_{i}
\end{equation}
\begin{proof}
    We can rewrite the update as a sum over all the data, where the probability of inclusion cancels out with the aggregation weight:
    \begin{align*}
        & \E_{\mathcal{K}} [\Delta^{t}] = \E_{\mathcal{K}} \left[\sum_{i \in \mathcal{C}} \frac{S}{C} p_i  \Delta^{t}_i + \sum_{i \in \mathcal{R}} \frac{N-S}{K-C} p_i \Delta^{t}_i\right] \\
        &= \E_{\mathcal{K}} \left[\sum_{i \in \mathcal{S}} \1_{\{i\in \mathcal{C}\}}\frac{S}{C} p_i  \Delta^{t}_i + \sum_{i \in \mathcal{N} \setminus \mathcal{S}} \1_{\{i\in \mathcal{R}\}}\frac{N-S}{K-C} p_i \Delta^{t}_i\right] \\
        &= \sum_{i \in \mathcal{S}} \frac{C}{S}\frac{S}{C}p_i \Delta^{t}_i +  \sum_{i \in \mathcal{N} \setminus \mathcal{S}} \frac{K-C}{N-S}\frac{N-S}{K-C}p_i \Delta^{t}_i \\ 
        &= \sum_{i=1}^N p_i \Delta^{t}_i
    \end{align*}
    where $\1_{\{\textrm{predictate}\}}$ is the indicator function with value $1$ when the predicate is true, and $0$ otherwise.
\end{proof}    
\end{theorem}
\normalsize
\Cref{sec:proof_of_convergence} shows that estimating unbiased updates is key to analyzing the convergence of \mytitle, following proof techniques from \citet{chen2020optimal,fraboni2021clustered}.

\subsection{Mask Shifting} 
\label{sec:mask-shiftnig}

Sticky sampling allows clients in a sticky group to be sampled more frequently. 
%
%
However, sticky sampling alone is insufficient. As we have seen in \Cref{fig:exp_prelim}, a client re-sampled after 10 rounds still needs to download around 50\%-80\% of the global model on average.
This is because the masked updates of two successive rounds (e.g., $\Tilde{\Delta}^t$ and $\Tilde{\Delta}^{t+1}$) have little overlap.

\begin{figure}[t]
    \centering
    \includegraphics[width=\linewidth]{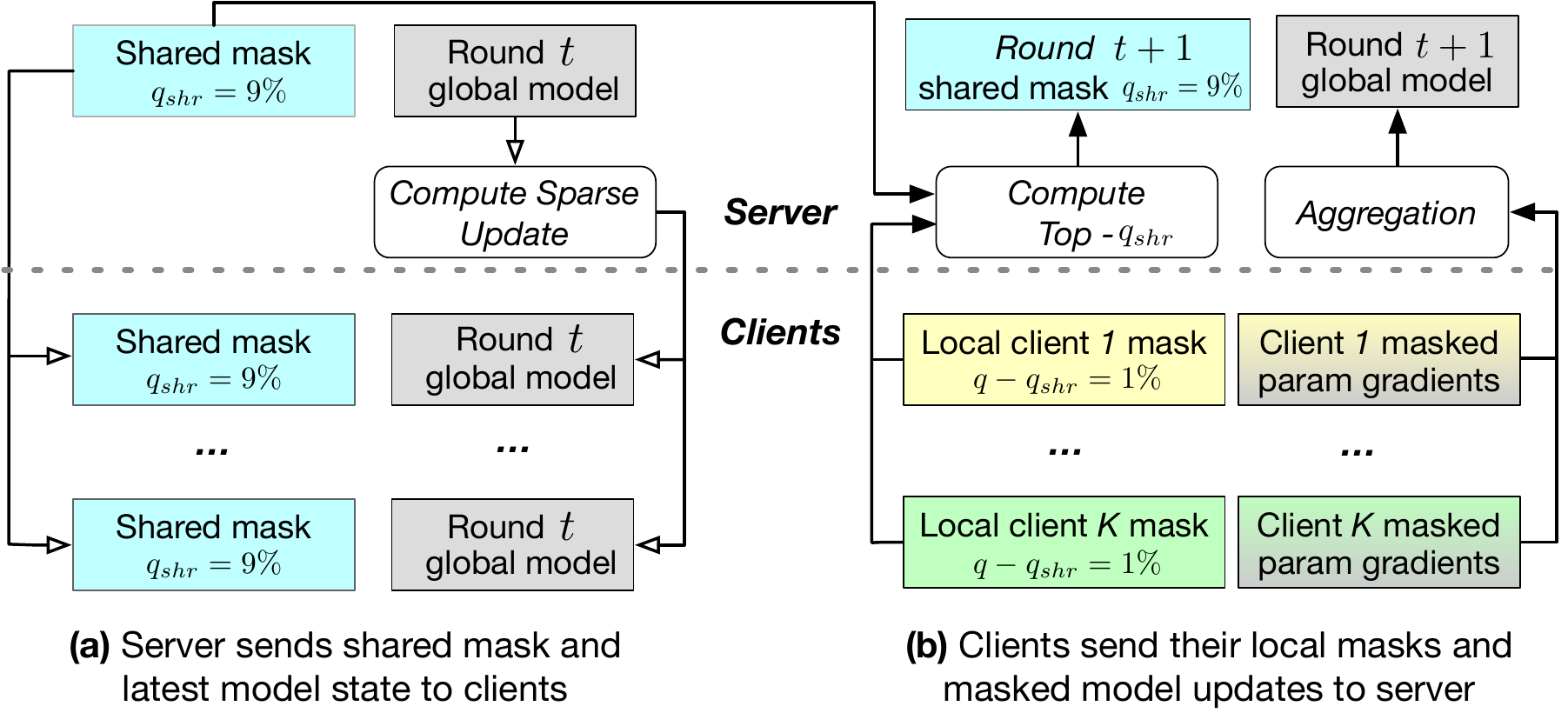}
    \caption{Mask shifting design with $q=10\%$ and $q_{shr}=9\%$. 
        }
    \label{fig:mask-shifting}
\end{figure}

We solve this issue by designing a gradual mask shifting strategy, that prevents the mask from changing too quickly while ensuring that the total compression ratio is maintained. \Cref{fig:mask-shifting} illustrates our mask shifting design. We construct a shared mask with compression ratio $q_{shr}$ (with $q_{shr} < q$), which is represented using a bitmap shared with selected clients in $M^t \in \mathbb{B}^d$ in round $t$. Clients send their update for parameters in $M^t$, as well as a $q - q_{shr}$ proportion of locally important parameters. The server will use $M^t$ as well as locally important parameters to calculate the model update, and to shift $M^t$ to obtain $M^{t+1}$, while keeping a large overlap between consecutive masks.

\Cref{algo:glueFL} details the mechanism, with sticky sampling from \Cref{algo:sticky_sampling} used to select clients in lines~\ref{glueFL-line:sticky_sampling} and~\ref{glueFL-line:sticky_sampling_update}. The server first synchronizes the global model $\mathbf{w}^t$ with sampled clients and distributes $M^t$ to them (line~\ref{glueFL-line:share_mask}). In line~\ref{glueFL-line:shared_gradients}, the client $i$ calculates the shared local gradient $\Tilde{\Delta}_{i, shr}^{t}$ as $M^t \odot \Tilde{\Delta}^t_i$, where $\odot$ sets those positions that are not covered by the masks to zero. Next, the algorithm computes unique local gradients $\Tilde{\Delta}_{i, uni}^{t}$ by selecting a $(q-q_{shr})$ proportion of the largest values in other (previously masked) positions, to provide more local information to the server (line~\ref{glueFL-line:unique_gradients}). Finally, client $i$ sends $\Tilde{\Delta}_{i, shr}^{t}$ and $\Tilde{\Delta}_{i, uni}^{t}$ to the server. 

During aggregation, the central server uses sticky sampling importance weights $\nu^t_{i,s}, \nu^t_{i,r}$ given in \S\ref{sec:sticky_sampling}.
The server first computes the shared update $\Tilde{\Delta}_{shr}^{t}$ based on all client (weighted) updates, and the update based on unique local information by selecting the $(q-q_{shr})$ proportion of largest overall (weighted) gradients (line~\ref{glueFL-line:update_computation}). Formally, each quantity is computed as:
\begin{gather}
    \Tilde{\Delta}_{shr}^{t} \gets \sum_{i \in \mathcal{C}} \nu_{i, s}^t \Tilde{\Delta}^{t}_{i,shr} + \sum_{i \in \mathcal{R}} \nu_{i, r}^t \Tilde{\Delta}^{t}_{i,shr}  \label{eq:shared_mask_aggregation} \\
    \Tilde{\Delta}_{uni}^{t} \gets top_{(q-q_{shr})} \left(\sum_{i \in \mathcal{C}} \nu_{i, s}^t \Tilde{\Delta}^{t}_{i,uni} + \sum_{i \in \mathcal{R}} \nu_{i, r}^t \Tilde{\Delta}^{t}_{i,uni}\right) \label{eq:unique_mask_aggregation}
\end{gather}
These updates are combined and update the global model (line~\ref{glueFL-line:model_update}).
Finally, the shared mask is updated by selecting a share $q_{shr}$ of parameters with the largest update values in the combined update (line~\ref{glueFL-line:mask_update}).
Since the new mask $M^{t+1}$ will be used to compute $\Tilde{\Delta}^{t+1}$, the overlap of two successive model updates $\Tilde{\Delta}^t$ and $\Tilde{\Delta}^{t+1}$ is at least $q_{shr}$. 
%

\setlength{\textfloatsep}{0.1cm}
\begin{algorithm2e}[t]
\footnotesize
\SetKwInOut{output}{Output}
\output{$\mathbf{w}^T$}
\For{$t \leftarrow 1$ \KwTo $T$}{
\CommentSty{/* Server:sticky sampling */} \\
Randomly select $|\mathcal{C}| = C$ clients from $\mathcal{S}$\; 
Randomly select $|\mathcal{R}| = K - C$ clients from $\mathcal{N} \setminus \mathcal{S}$\; 
Set of sampled clients $\mathcal{K} \gets \mathcal{C} \cup \mathcal{R}$ \label{glueFL-line:sticky_sampling} \;
Synchronize $\mathbf{w}^t$ with $\mathcal{K}$ by sending model updates\;
Send shared mask $M^t$ to $i \in \mathcal{K}$\; \label{glueFL-line:share_mask}
\CommentSty{/* Client:local training */}\\
\For{$i \in \mathcal{K}$ in parallel }{
$\mathbf{w}^{t,0}_i \gets \mathbf{w}^{t}$ \;
    \For{$e \leftarrow 0$ \KwTo $E - 1$}{
        $\mathbf{w}^{t,e+1}_i \gets \mathbf{w}^{t,e}_i - \gamma \mathbf{g}^{t,e}_i$ \;
    }
\CommentSty{/* Client:masking */} \\
$\Delta_i^{t} \gets \mathbf{w}^{t,E}_i - \mathbf{w}^{t,0}_i$ \;
$\Tilde{\Delta}_{i, shr}^{t} \gets M^t \odot \Delta^t_i$ \; \label{glueFL-line:shared_gradients}
$\Tilde{\Delta}_{i, uni}^{t} \gets top_{(q-q_{shr})}(\neg M^t \odot \Delta^t_i) $ \; \label{glueFL-line:unique_gradients}
}
\CommentSty{/* Server:aggregation */} \\
Receive $\Tilde{\Delta}_{i, shr}^t, \Tilde{\Delta}_{i, uni}^t$ from worker $i \in \mathcal{K}$ \;
Compute $\Tilde{\Delta}_{shr}^{t}$ via \Cref{eq:shared_mask_aggregation} \;
Compute $\Tilde{\Delta}_{uni}^{t}$ via \Cref{eq:unique_mask_aggregation} \;
$\Tilde{\Delta}^{t} \gets \Tilde{\Delta}_{shr}^{t} + \Tilde{\Delta}_{uni}^{t}$\; 
\label{glueFL-line:update_computation}
$\mathbf{w}^{t+1} \gets \mathbf{w}^{t} + \Tilde{\Delta}^{t}$\; \label{glueFL-line:model_update}
\CommentSty{/* Server:update shared mask */} \\
$M^{t+1} \gets top_{q_{shr}}(\Tilde{\Delta}_{shr}^{t} + \Tilde{\Delta}_{uni}^{t})$ \; \label{glueFL-line:mask_update}
\CommentSty{/* Server:update sticky group $\mathcal{S}$ */} \\ 
Randomly remove $K - C$ clients in $\mathcal{S} \setminus \mathcal{C}$ \; \label{glueFL-line:sticky_sampling_update}
$\mathcal{S} \gets \mathcal{S} \cup \mathcal{R}$ \;
}

\caption{\mytitle}
\label{algo:glueFL}

\end{algorithm2e}
\normalsize

\subsection{Adapting other Techniques to Work with \mytitle}
\label{sec:other-tweaks}

We further improve the performance of \mytitle by adapting common FL  techniques to sticky sampling and mask shifting~\cite{chen2021communication, han2020adaptive, tang2019doublesqueeze,qian2021error, wu2018error, seide20141}.

\paragraph{Shared Mask Regeneration}
Previous work~\cite{chen2021communication, han2020adaptive} showed that model parameters converge at different rates. Meanwhile, a parameter that has converged may become unstable in later rounds. 
%
For example, according to \citet{chen2021communication}, it is possible that some 10\% of parameters are unstable in both round $t$ and $t+1$, while another 5\% parameters are only unstable in round $t+1$. In this case, a small $(q-q_{shr})$ (e.g., $2\%$) value will slow down convergence, as the shared mask fails to cover the gradients of the unstable $5\%$ of parameters and a large  $(q-q_{shr})$ (e.g., $10\%$) value incurs more bandwidth cost.
%

To address this, we use a small $(q-q_{shr})$ value while re-generating the entire shared mask $M^t$ every $I$ rounds. To regenerate, we set $q_{shr}=0$ and update $M^t$ as $top_{q_{shr}}(\Tilde{\Delta}_{uni}^t)$ (\Cref{algo:glueFL}, line~\ref{glueFL-line:mask_update}). 
%
%
Although this process introduces more downstream overhead in the next few rounds, it speeds up training and reduces overall bandwidth.
%

\paragraph{Error-Compensation}
Compression methods, such as quantization and sparsification, slow down model convergence due to the loss of information in client updates~\cite{tang2019doublesqueeze,qian2021error,wu2018error}. Error-compensation is a technique to alleviate this problem, first proposed to accelerate convergence in 1-bit SGD~\cite{seide20141}. The key idea is for clients to (1) remember their local compression error (the difference between their true update and what is actually sent to the server), and (2) add it into the next round's computed local gradient before compression. 
In \mytitle, we apply error compensation as:
\begin{equation} \label{eq:compensation}
    \Delta_i^t \gets \Delta_i^t + \frac{\nu_i^{\varphi(t)}}{\nu_i^t} \cdot h_i^{\varphi(t)}
\end{equation}
where $\nu_i^{t}$ is the aggregation weight applied at step $t$ for client $i$ (i.e., $\nu_{i,s}^t$ if they are in the sticky group, $\nu_{i,r}^t$ otherwise; their exact values are defined in \S\ref{sec:sticky_sampling}), $\varphi(t)$ indicates the step-index when client $i$ was last selected, and $h_i^{\varphi(t)}$ the compensation vector for client $i$ in round $\varphi(t)$. After that, the client computes $\Tilde{\Delta}_{i, shr}^t$ and $\Tilde{\Delta}_{i, uni}^t$ (\Cref{algo:glueFL}, lines~\ref{glueFL-line:shared_gradients}-\ref{glueFL-line:unique_gradients}). Then, the compensation vector is calculated as $h_i^{t} = \Delta_i^t - (\Tilde{\Delta}_{i, shr}^t + \Tilde{\Delta}_{i, uni}^t)$. 

The reason for scaling with $h_i^{\varphi(t)}$ in \Cref{eq:compensation} is to ensure that client $i$'s compensation is consistent with the aggregation in sticky sampling. 
As the compensation only applies to a client's local gradient before masking, this optimization does not introduce extra bandwidth and improves convergence performance. 


\section{Convergence Analysis}
\label{sec:convergence}
From a theoretical perspective, we show that \mytitle without masking can achieve convergence at a rate of $O(1/\sqrt{T})$ for smooth non-convex functions under two assumptions (\S\ref{subsec:assumption}).
\S\ref{subsec:convergence} states our result and their interpretation, with details in \S\ref{sec:proof_of_convergence}.

\subsection{Assumptions} \label{subsec:assumption}
\newcommand{\R}{\mathbb{R}}

We make a standard assumption that clients sample a mini-batch in each local update such that the computed gradient is equal to the true gradient in expectation~\cite{karimireddy2020scaffold, li2019convergence, wang2020tackling, yang2021achieving, wu2022sign}. That is, $\E_{\xi_i \sim \mathcal{D}_i} \nabla f_i(\mathbf{w}, \xi_i) = \nabla F_i(\mathbf{w})$ for all workers $i \in \{1, \dots, N\}$ and the model $\w \in \R^d$, where $\xi_i$ and $\mathcal{D}_i$ represent the mini-batch and the local training set, respectively. 
%
We make two more assumptions:

\begin{assumption}[Bounded Local Variance] \label{ass:bounded_var}
There exists a constant $\sigma>0$, such that the variance of each local gradient estimator is bounded by,
\begin{equation*}
    \mathbb{E}_{\xi_i \sim \mathcal{D}_i} \|\nabla f_i(\mathbf{w}, \xi_i) - \nabla F_i(\mathbf{w})\| \leq \sigma^2, \quad \forall i \in [N].
\end{equation*}
\end{assumption}

We also
assume that the local objective functions (i.e., $F_1, \dots, F_N$) and their derivatives are Lipschitz continuous. 


\begin{assumption}[Continuity and Smoothness] \label{ass:L-smooth}
The local objective functions are $L_c$-continuous and $L_s$-smooth.
\end{assumption}

\subsection{Convergence Result} \label{subsec:convergence}
Here we analyze the convergence rate of sticky sampling (\Cref{algo:sticky_sampling}) on non-convex local objective functions. See \S\ref{sec:proof_of_convergence} for the complete proof. 
\begin{theorem}
\label{theorem:convergence}
Suppose \Cref{ass:L-smooth,ass:bounded_var} hold, and set the aggregation weights as $\nu^t_{i,s} = \frac{S}{C} p_i$ and $\nu^t_{i,r} = \frac{N-S}{K-C} p_i$. Let the learning rate be
\begin{align}
    \gamma =  \sqrt{\frac{1}{E (\sigma^2 + E)} \cdot \frac{K}{TA}}
\end{align}
Algorithm \ref{algo:sticky_sampling} is such that:
\footnotesize
\begin{align}
    \min_{t\in\{1, \dots, T\}} \norm{\nabla F(\w^t)} = O\left(\sqrt{\left(1 + \frac{\sigma^2}{E}\right) \cdot \frac{A}{KT}}\right) + O\left(\frac{K}{TA}\right)
\end{align}
\normalsize
where $A=\frac{K}{N}\left(\frac{S^2}{C} + \frac{(N-S)^2}{K-C}\right) \left(\sum_{i=1}^N p_i^2\right)$. We treat $L_s$, $L_c$, and $F(\w^1)-F^*$ as constants.
\end{theorem}
This result gives a convergence rate for reaching a fixed point during model training.

\textbf{Comparison with FedAvg.}
If all clients have equal weights, (i.e., $p_i = \frac{1}{N}$ for all workers $i \in \{1, \dots, N\}$), and the sticky group does not exist (i.e., $S=0$), the algorithm reduces to FedAvg, and $A = 1$. As we can see, when we set the number of local updates $E \geq \sigma^2$ and $T$ is sufficiently large, the convergence result is led by $O\left(\sqrt{\frac{1}{KT}}\right)$. This is comparable to the state-of-the-art works on convergence of FedAvg as described in \S\ref{sec:background_fl}.
Sticky-sampling introduces a variance cost (the $\frac{S^2}{C} + \frac{(N-S)^2}{K-C}$ term in $A$) to remain unbiased under non-uniform client sampling. Next, we show empirically that this is a favorable trade-off given the bandwidth savings enabled by sticky sampling (\S\ref{sec:experiment}).

\section{Experimental Evaluation} 
\label{sec:experiment}
We evaluate \mytitle across several datasets and network distributions.
%
Our goal is to answer three questions:
\begin{enumerate}[label=Q\arabic*:]
    \setlength{\itemsep}{0pt}
    \setlength{\parsep}{0pt}
    \setlength{\parskip}{0pt}
    \item What model accuracy does \mytitle achieve?
    \item How does \mytitle impact bandwidth usage?
    \item How quickly does the model converge with \mytitle?
\end{enumerate}

\subsection{Experimental Setup}
\label{sec:experimental_setup}
We deployed \mytitle on a set of VMs in one data-center with a total of 14 NVIDIA Tesla V100 GPUs. To reproduce real-world heterogeneous client performance, we use FedScale's client behavior trace and the NDT dataset~\cite{mlab} to simulate the availability pattern and bandwidth capacity of clients, respectively. To mitigate stragglers and offline clients, FedScale introduces an \texttt{over-commitment (OC)} variable~\cite{bonawitz2019towards} which we set to 1.3 in all experiments. That is, we sample $1.3 \times K$ clients in each round and use the first $K$ uploaded updates. 

\paragraph{Datasets and Models} We use three datasets: FEMNIST~\cite{caldas2018leaf}, OpenImage~\cite{OpenImages}, Google Speech~\cite{warden2018speech}. The first two datasets are frequently used for image classification and consist of 640K and 1.3M colored images, respectively. Google Speech is a dataset with 105K speech samples. We partition the data using FedScale's real-world non-iid client-data mapping~\cite{fedscale-icml22} and remove those clients that have fewer than 22 samples as the default setting in FedScale. In total, we use $2,800$, $10,625$, and $2,066$ clients in our
experiments, respectively. 
The models we use are ShuffleNet~\cite{zhang2018shufflenet} and MobileNet~\cite{sandler2018mobilenetv2} for both FEMNIST and OpenImage, and ResNet-34~\cite{he2016deep} for Google Speech. Batch normalization layers in these models must be treated differently because some related parameters are non-trainable. We describe our approach for this in \Cref{sec:bn}. 
We set the number of sampled clients $K=30$, $100$, and $30$ for FEMNIST, OpenImage, and Google Speech, respectively. 

%

\paragraph{Baselines} We compare \mytitle with FedAvg~\cite{mcmahan2017communication}, the most widely used FL algorithm with no model compression methods. We also compare \mytitle with STC~\cite{sattler2019robust} and APF~\cite{chen2021communication}, which are the state-of-the-art sparsification and parameter freezing strategies, respectively.


\paragraph{Metrics} We measure the total data volume and total training time to address Q2 and Q3, respectively. We also analyze the downstream bandwidth and download time. For download time, we pick the slowest client in each round and sum up their download time. To address Q1, similar to Oort~\cite{oort-osdi21}, we average the test accuracy over 5 rounds and report the results when the averaged accuracy first reaches the target accuracy. 


\begin{table*}[!ht] 
\centering 

\renewcommand{\arraystretch}{1.3}
\normalsize
\caption{\textbf{D}ownstream transmission \textbf{V}olume (\textbf{DV}, in $\times10^2$ GB) and \textbf{D}ownload \textbf{T}ime (\textbf{DT}, in hours) for training different models/datasets. We also present \textbf{T}otal transmission \textbf{V}olume (\textbf{TV}) and \textbf{T}otal training \textbf{T}ime (\textbf{TT}) in parenthesis. We measure Top-1 accuracy for FEMNIST and Google Speech, and Top-5 accuracy for OpenImage \cite{fedscale-icml22}. We set the target accuracy to be the highest achievable accuracy by all approaches. Our target accuracies are comparable with previous work~\cite{oort-osdi21, fedscale-icml22}. The best results are in bold.
}
\scriptsize

\begin{tabular}{w{c}{14mm}w{c}{8.5mm}w{c}{8mm}w{c}{9mm}w{c}{2mm}w{c}{4mm}|w{c}{2mm}w{c}{4mm}||w{c}{2mm}w{c}{4mm}|w{c}{2mm}w{c}{4mm}|w{c}{2mm}w{c}{4mm}|w{c}{2mm}w{c}{4mm}|w{c}{2mm}w{c}{4mm}|w{c}{2mm}w{c}{4mm}}

\hline 
\multirow{2}{*}{Dataset}

& \multirow{2}{*}{\# Clients}
& \multirow{2}{*}{\makecell{Target\\Acc.}}

& \multirow{2}{*}{Model} 
& \multicolumn{4}{w{c}{6mm}||}{\textbf{FedAvg}
}
& \multicolumn{4}{w{c}{6mm}|}{\textbf{STC}
}
& \multicolumn{4}{w{c}{6mm}|}{\textbf{APF}
}
& \multicolumn{4}{w{c}{6mm}}{\textbf{\mytitle (ours)}}

\\

\cline{5-20}
& & 
 &   & \textbf{DV} & (\textbf{TV}) & \textbf{DT} & (\textbf{TT}) & \textbf{DV} & (\textbf{TV}) & \textbf{DT} & (\textbf{TT})
 & \textbf{DV} & (\textbf{TV}) & \textbf{DT} & (\textbf{TT}) & \textbf{DV} & (\textbf{TV}) & \textbf{DT} & (\textbf{TT})\\


\hline
 

 \multirow{2}{*}{FEMNIST} 
 &  \multirow{2}{*}{2,800} 
 &  \multirow{2}{*}{73.3\%} & ShuffleNet & 2.6 & (4.6) &  2.7 & (7.6) & 2.6 & (3.4) & 2.7 & (5.7) & 2.3 & (3.2) & 2.3 & (5.7) & \textbf{2.2} & \textbf{(3.1)} & \textbf{2.2} &  \textbf{(5.3)}\\
 \cline{4-20}
 & & 
 &  MobileNet & 1.2 & (2.1) & 1.5 & (4.6) & 1.5 & (1.9) & 1.7 & (3.9) & 1.5 & (2.0) & 1.6 & (4.5) & \textbf{0.9} & \textbf{(1.4)} & \textbf{0.8} & \textbf{(3.3)} \\
 
 \hline
\multirow{2}{*}{OpenImage} 
&  \multirow{2}{*}{10,625} 
& \multirow{2}{*}{66.8\%} & ShuffleNet & 25.2 & (45.0) &   11.2 & (28.8) & 33.9 & (50.0) & 14.8 & (29.9) &  27.1 & (43.1) &  12.3 & (29.8) & \textbf{21.3} & \textbf{(31.4)} & \textbf{8.0} & \textbf{(19.2)}\\
 \cline{4-20}
 & 
 & 
 & MobileNet & 17.4 & (31.1) & 7.1 & (22.4) & 16.7 & (24.5) &  7.1 & (19.1) & 20.3 & (30.9) & 8.8 &  (21.0) & \textbf{14.9} & \textbf{(22.1)} & \textbf{5.8} & \textbf{(14.4)} \\
 \hline
 Google Speech 
 & 2,066 
 & 61.2\%  & ResNet-34 & 12.8 & (23.0) & 20.1 & (60.9) & 13.5 & (18.5) & 16.0 & (42.3) & 15.8 & (21.9) & 19.1 & (54.1) & \textbf{7.2} & \textbf{(12.5)} & \textbf{12.1} & \textbf{(27.8)}\\
\hline
\end{tabular}
\vspace{-15pt}
\label{table:accuracy}
\end{table*}

\normalsize
\paragraph{Training Parameters}
Clients perform $10$ local updates per round. 
We use PyTorch's SGD optimizer with a momentum factor of 0.9 for all tasks. For FEMNIST, OpenImage, and Google Speech, the initial learning rate is set to 0.01, 0.05, and 0.01, respectively, with a decay factor of 0.98 every 10 rounds. To obtain the best performance, we set the total mask ratio $q=20\%$ for ShuffleNet, and $q=30\%$ for MobileNet and ResNet-34 in STC. For APF, we set the threshold for effective perturbation, which reflects the compression ratio, to $0.1$ for all tasks. The remaining STC and APF parameters are set to their optimal values~\cite{sattler2019robust, chen2021communication}. For \mytitle, the default sticky group parameters are $S=4K$ and $C=4K/5$.  For ShuffleNet, the default mask shifting parameters are $q=20\%$ and $q_{shr}=16\%$. For MobileNet and ResNet-34, we set $q=30\%$ and $q_{shr}=24\%$. We use $I=10$ to regenerate the shared mask every 10 rounds. We choose these values as they produce the best performance across most tasks.

\subsection{Performance Results}
\label{sec:performance}


\paragraph{Communication costs}

\Cref{table:accuracy} lists the data volume and training time for FedAvg (baseline), STC, APF, and \mytitle (our framework). It shows that STC and APF outperform FedAvg as they require less bandwidth to reach the target accuracy, reducing volume by $8\%$ on average. However, STC and APF consume substantial downstream bandwidth. For example, when training MobileNet on FEMNIST, STC only takes 40 GB to upload gradients but uses 150 GB for downstream synchronization.
\mytitle reduces downstream bandwidth (\Cref{table:accuracy}): for OpenImage, \mytitle provides a saving of 15\% compared with FedAvg, while for Google Speech \mytitle saves 42\%.
We further compare the performance of \mytitle with STC and APF.
In each case, while consuming nearly the same amount of upstream bandwidth (note upstream bandwidth volume = \textbf{TV}-\textbf{DV} in \Cref{table:accuracy}), \mytitle uses the least downstream bandwidth across all three datasets. For example, when training MobileNet on OpenImage, APF, STC, and \mytitle all consume around 900 GB to upload gradients. However \mytitle lowers download bandwidth by 11\% and 26\% as compared with STC and APF, respectively. 
This is because STC and APF do not bound the changes of masks in a communication round and the update size rapidly increases.

\paragraph{Wall-clock Time}
\Cref{table:accuracy} indicates that downstream bandwidth  
is the bottleneck. For example, when training MobileNet on FEMNIST, FedAvg uses 32\% of its total training time for model synchronization while STC uses 43\%. \mytitle reduces total training time by reducing downstream bandwidth and saving download time, which speeds up the training by 15\% and 26\% as compared with STC and APF.




\begin{figure}[!t]
\centering
\vspace{-10px}
\subfloat[ShuffleNet (FEMNIST)]{
\centering
\includegraphics[width=0.23\textwidth]{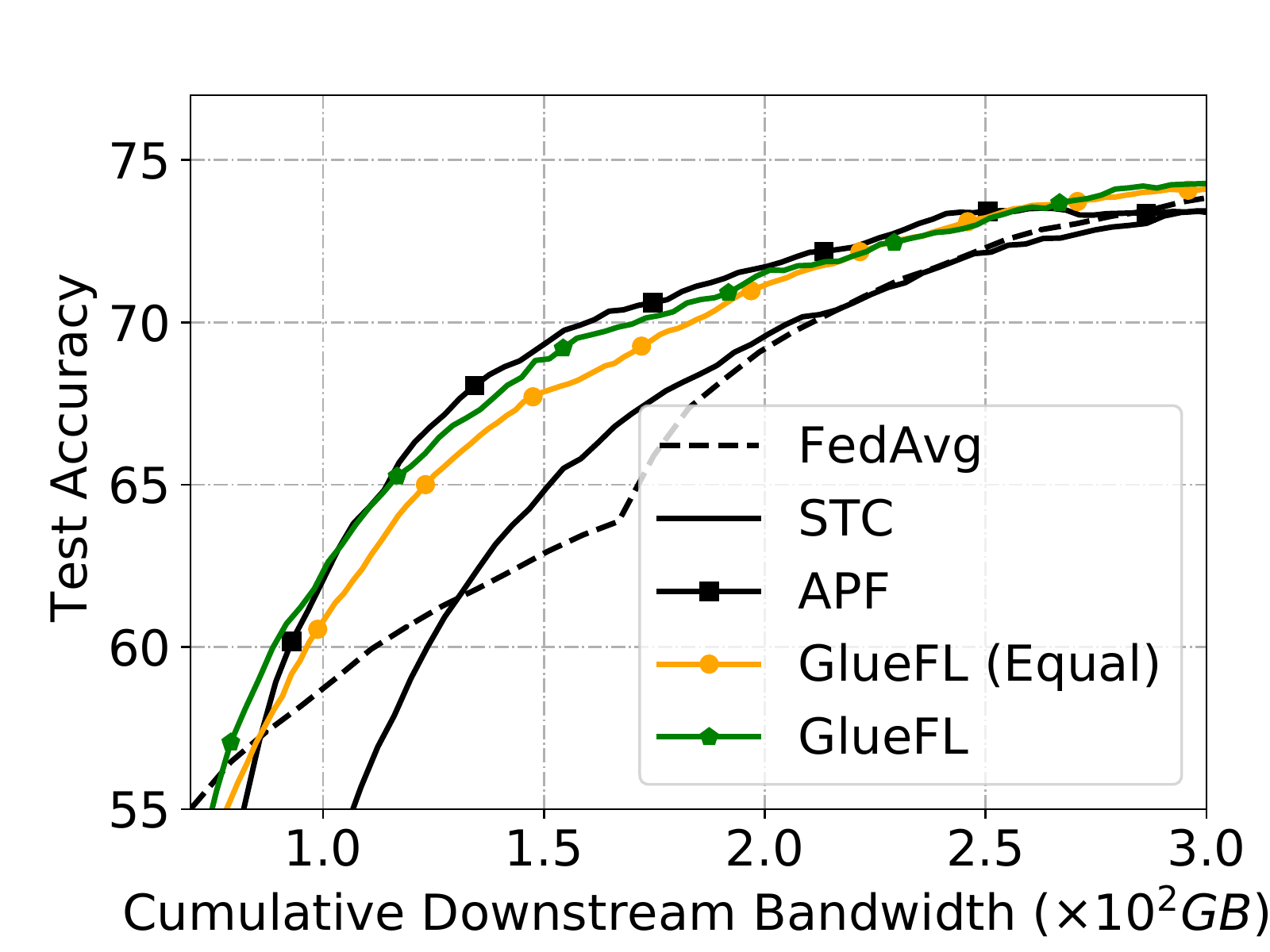}
}
\subfloat[ResNet-34 (Google Speech)]{
\centering
\includegraphics[width=0.23\textwidth]{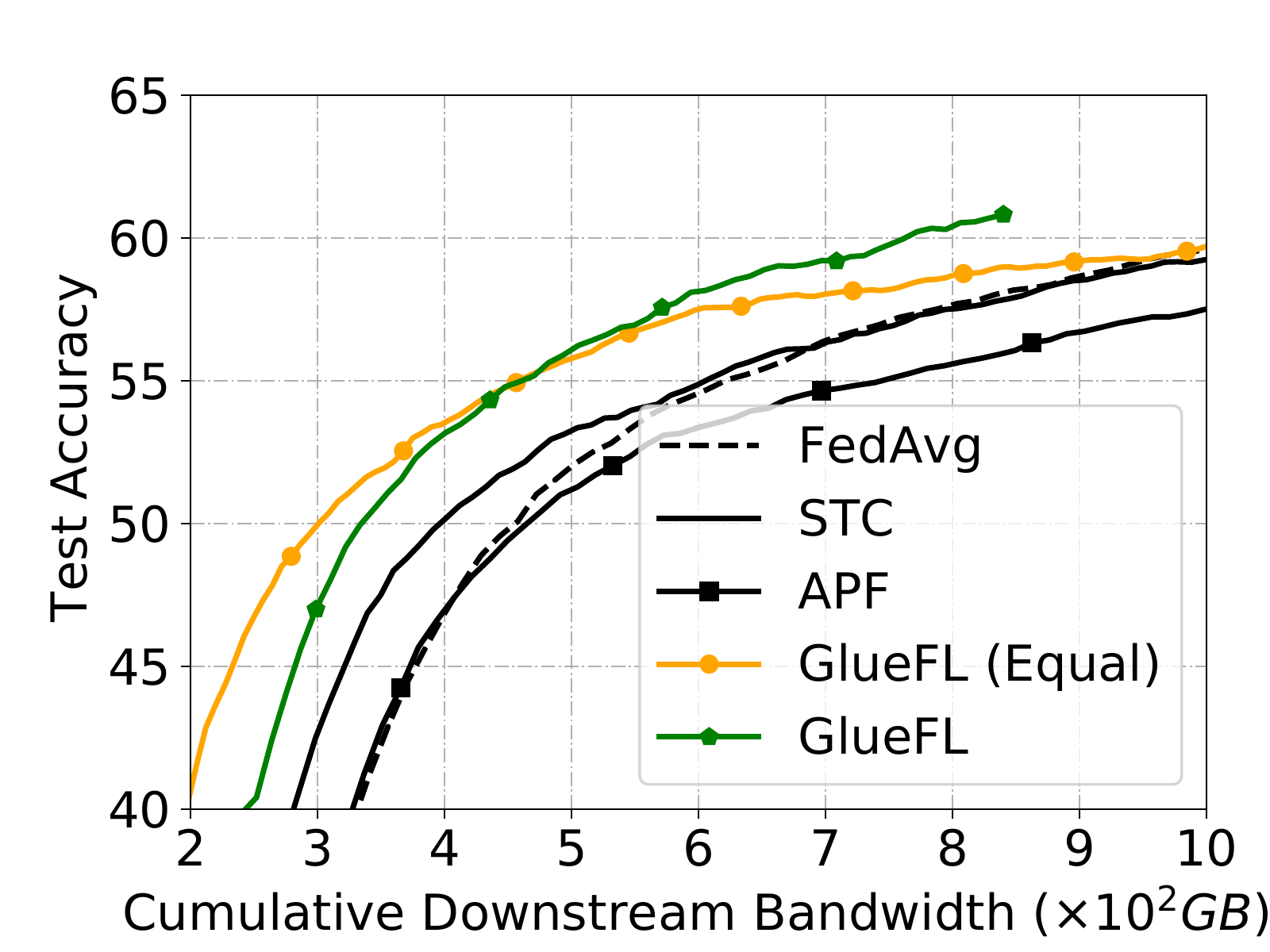}
}
\caption{Effect of aggregation weights $\nu^t_{i,s}$ and $\nu^t_{i,r}$: GlueFL (Equal) is biased (equal weights), while GlueFL is unbiased.
}
\label{fig:parameter_aggregation_weight}
\end{figure}

\begin{figure}[!t]
\centering
\vspace{-10px}
\subfloat[ShuffleNet (FEMNIST)]{
\centering
\includegraphics[width=0.23\textwidth]{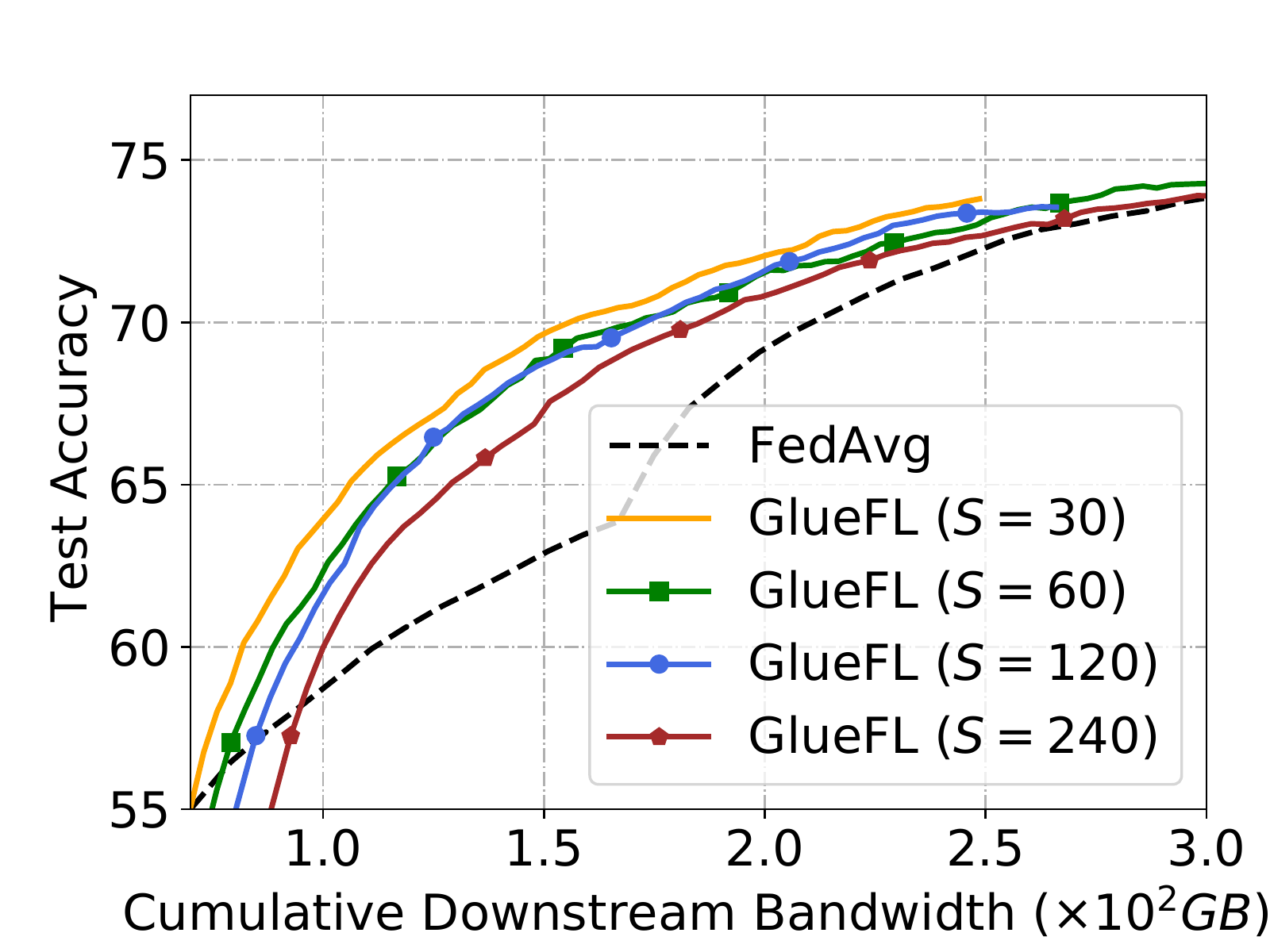}
}%
\subfloat[ResNet-34 (Google Speech)]{
\centering
\includegraphics[width=0.23\textwidth]{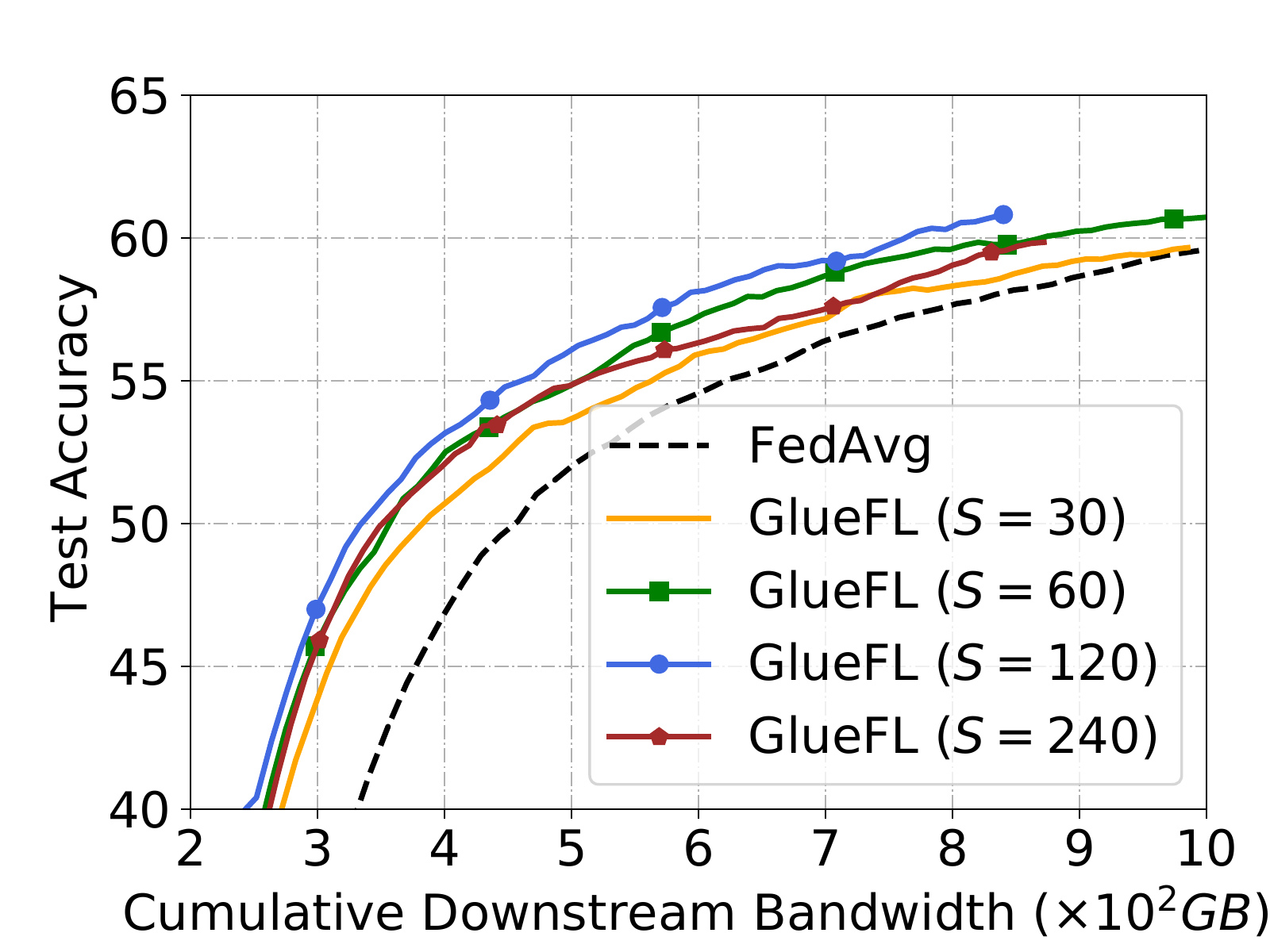}
}

\caption{Effect of sticky group size $S$.} 

\label{fig:parameter_sticky_group_size}
\end{figure}

\subsection{Sensitivity Analysis}
\label{sec:exp_para}
We evaluate the influence of \mytitle parameters on training performance on FEMNIST with ShuffleNet and Google Speech with ResNet-34. Similar to \S\ref{sec:performance}, we use $K=30$.
When evaluating one parameter, we use defaults for the others (see \S\ref{sec:experimental_setup}). For each setting, we run \mytitle for 1,000 rounds and report the average test accuracy over 20 rounds with respect to the cumulative downstream bandwidth.

\paragraph{Effect of aggregation weights $\nu^t_{i,s}$ and $\nu^t_{i,r}$} 
\Cref{fig:parameter_aggregation_weight} demonstrates the impact of two settings of aggregation weights on training performance: equal (i.e., $\nu^t_{i,s}=\nu^t_{i,r}=1/K$) and unbiased (see \S\ref{sec:sticky_sampling}). Overall, unbiased aggregation weights lead to similar or better convergence speed for the same amount of cumulative downstream bandwidth usage. In the case of Google Speech, unbiased aggregation was able to achieve convergence while saving 41\% of downstream bandwidth. 


\begin{figure}[!t]
\centering
\vspace{-10px}
\subfloat[ShuffleNet (FEMNIST)]{
\centering
\includegraphics[width=0.23\textwidth]{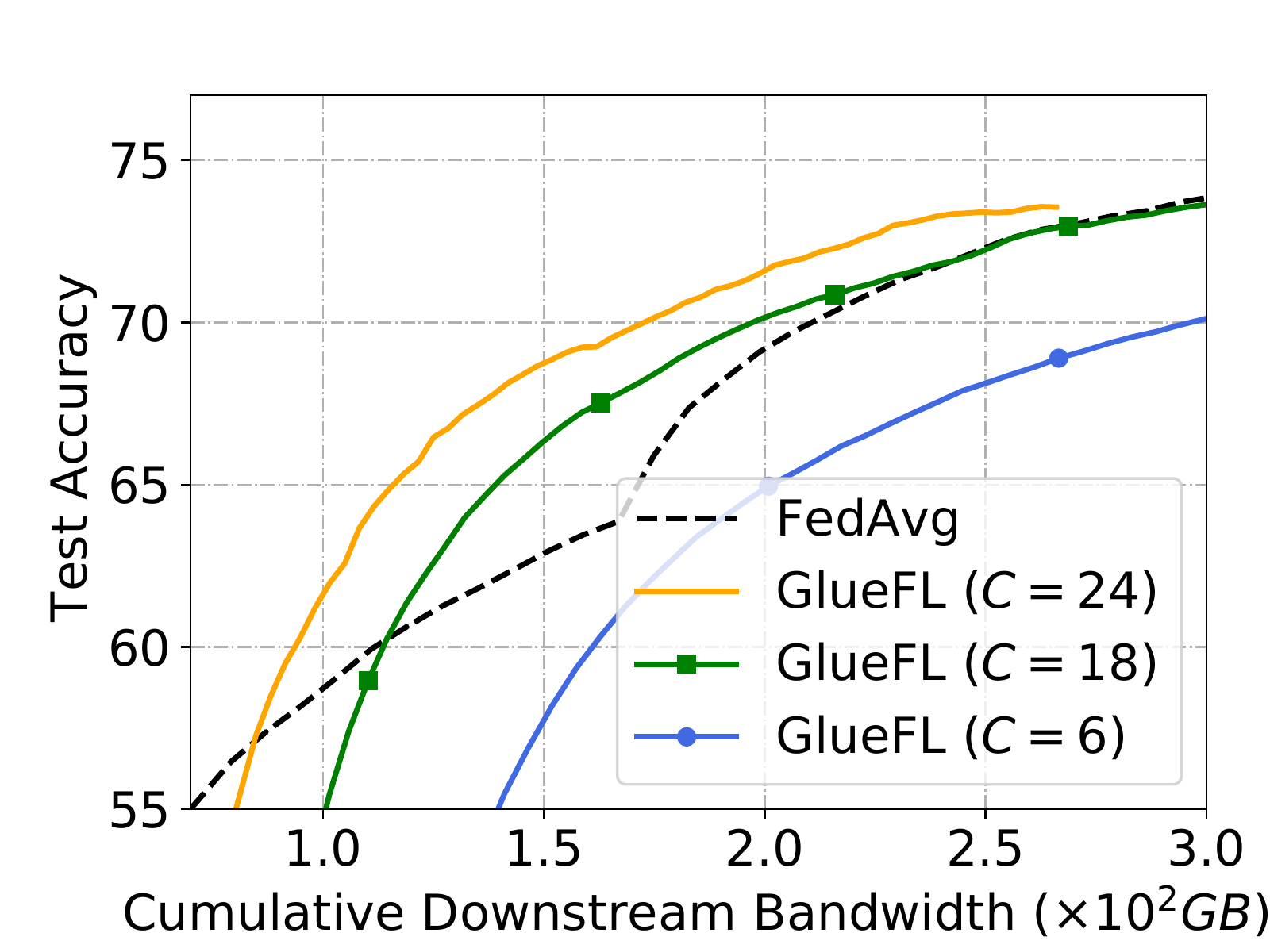}
}
\subfloat[ResNet-34 (Google Speech)]{
\centering
\includegraphics[width=0.23\textwidth]{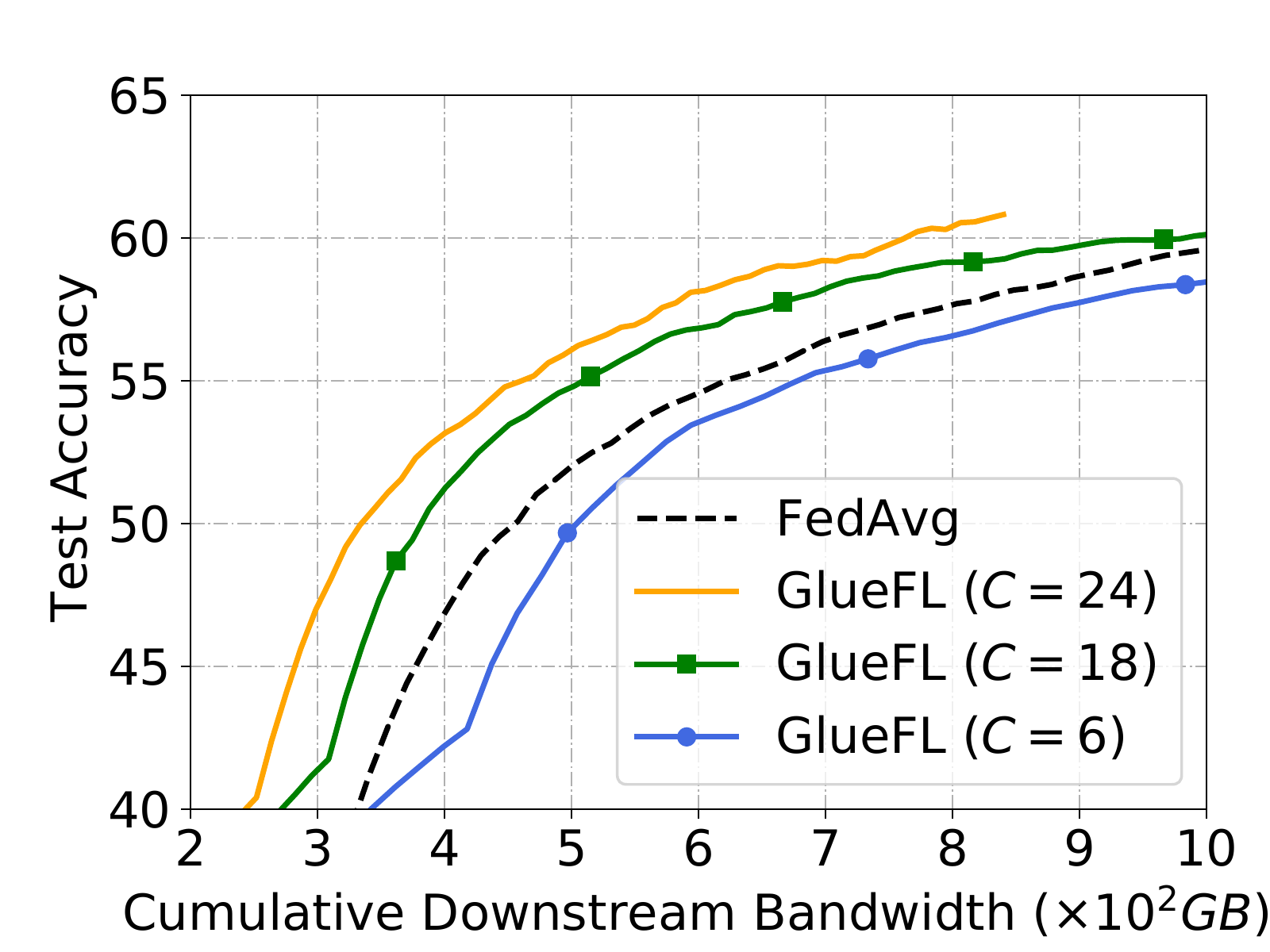}
}
\caption{Effect of sticky sampling parameter $C$. 
}
\label{fig:parameter_sticky_replacement_ratio}
\end{figure}

\begin{figure}[!t]
\centering
\vspace{-10px}
\subfloat[ShuffleNet (FEMNIST)]{
\centering
\includegraphics[width=0.23\textwidth]{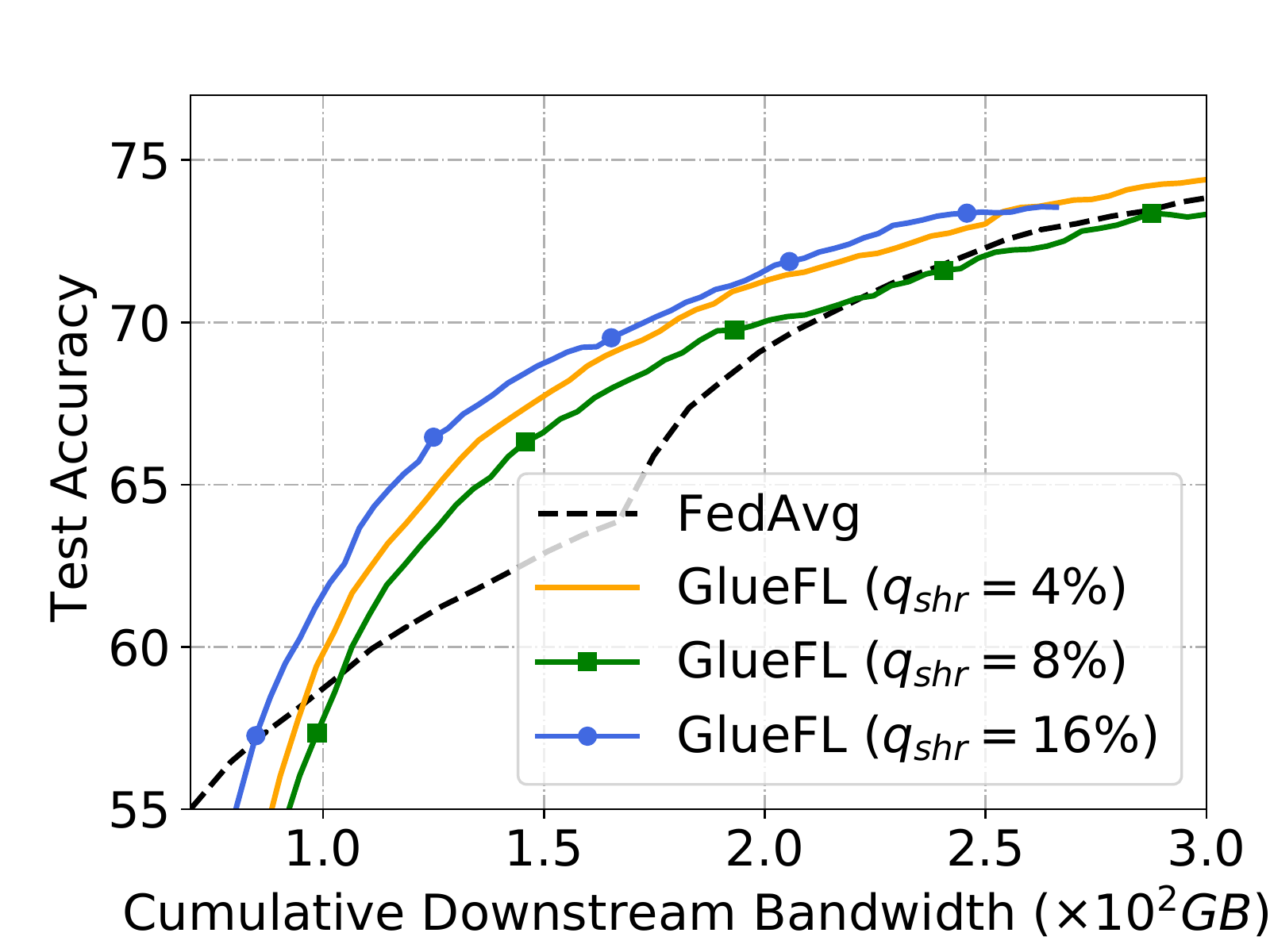}
}
\subfloat[ResNet-34 (Google Speech)]{
\centering
\includegraphics[width=0.23\textwidth]{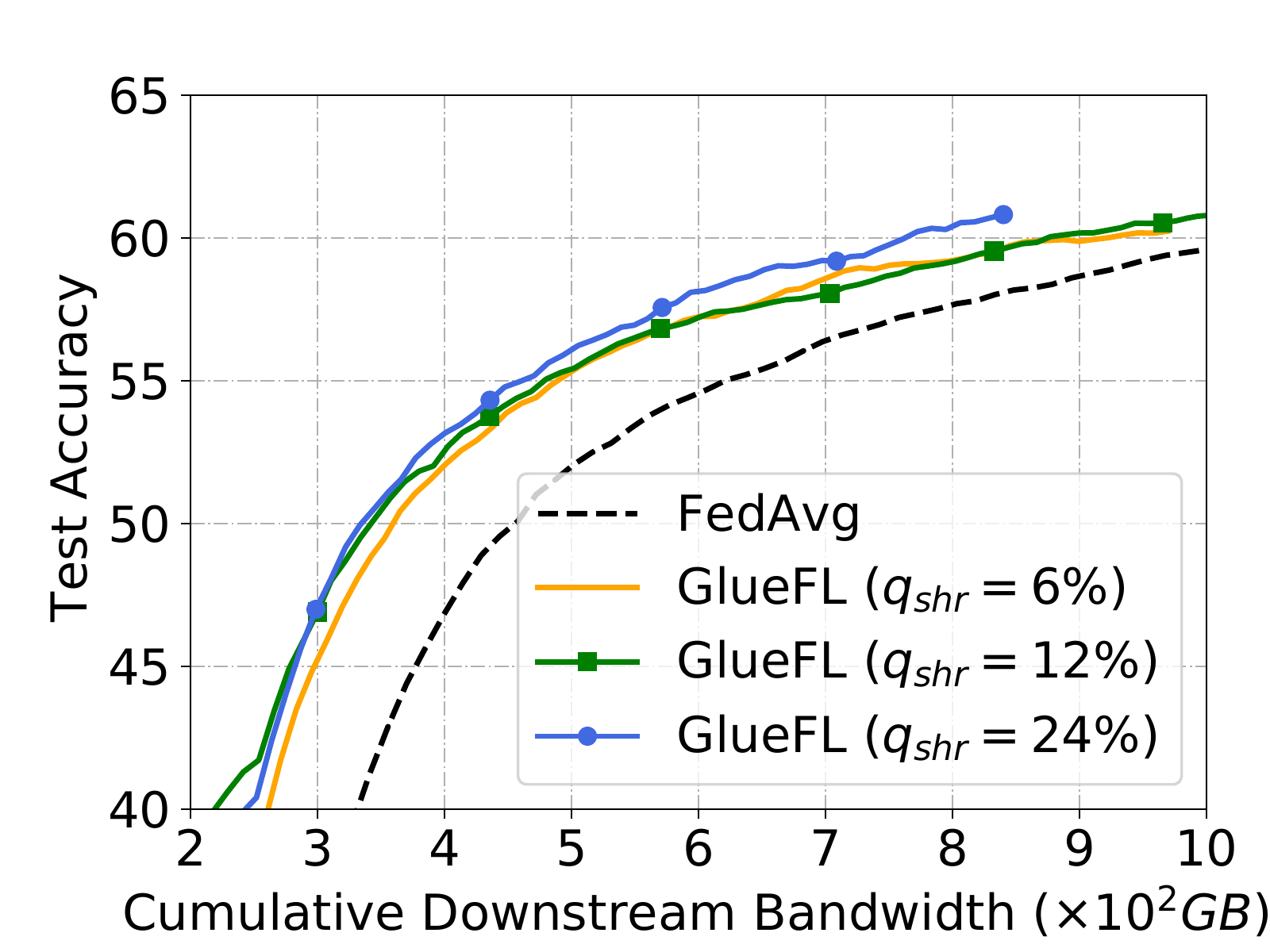}
}
\caption{Effect of shared mask ratio  $q_{shr}$.
}

\label{fig:parameter_shared_mask_ratio}
\end{figure}

\paragraph{Sticky sampling parameters $S$ and $C$}
\Cref{fig:parameter_sticky_group_size} shows the impact of sticky group size $S$ on training performance. 
Typically, a larger sticky group size means more diverse training data for the sticky clients and indirectly better accuracy at the cost of more communication. It follows that choosing an appropriately large $S$ is important for optimizing performance. For instance, the $S=120$ setting for Google Speech reached the target accuracy with almost 20\% less downstream communication compared with $S=60$. However, the same $S$ is unable to help \mytitle achieve a speedup for FEMNIST.


Next, we evaluate the impact of the sticky sampling parameter $C$ (\Cref{fig:parameter_sticky_replacement_ratio}). $C$ clients in the sticky group are sampled and $(K - C)$ clients are replaced by clients from the non-sticky group. 
Across $C=6, 18,\ \mathrm{and }\ 24$ in \Cref{fig:parameter_sticky_replacement_ratio}, we do not observe a large improvement in accuracy for smaller $C$. By contrast, $C=6$ adds 76\%  download bandwidth in each round as \mytitle is unable to capitalize on the savings from sticky sampling due to more new clients. This indicates that a large $C$ does not harm accuracy and saves more bandwidth.


\paragraph{Mask shifting parameter $q_{shr}$} \Cref{fig:parameter_shared_mask_ratio} shows the effect of the shared mask ratio $q_{shr}$ on performance. On average, a higher value ($q_{shr}=16\%$) does not cause accuracy to drop substantially and is preferable as \mytitle uses the least downstream bandwidth to reach the convergence accuracy of FedAvg. This is because \mytitle optimizes mask shifting with shared mask regeneration and error compensation. 

\begin{figure*}[!ht]
\centering
\subfloat[End-user device network]{
\centering
\includegraphics[width=0.29\textwidth]{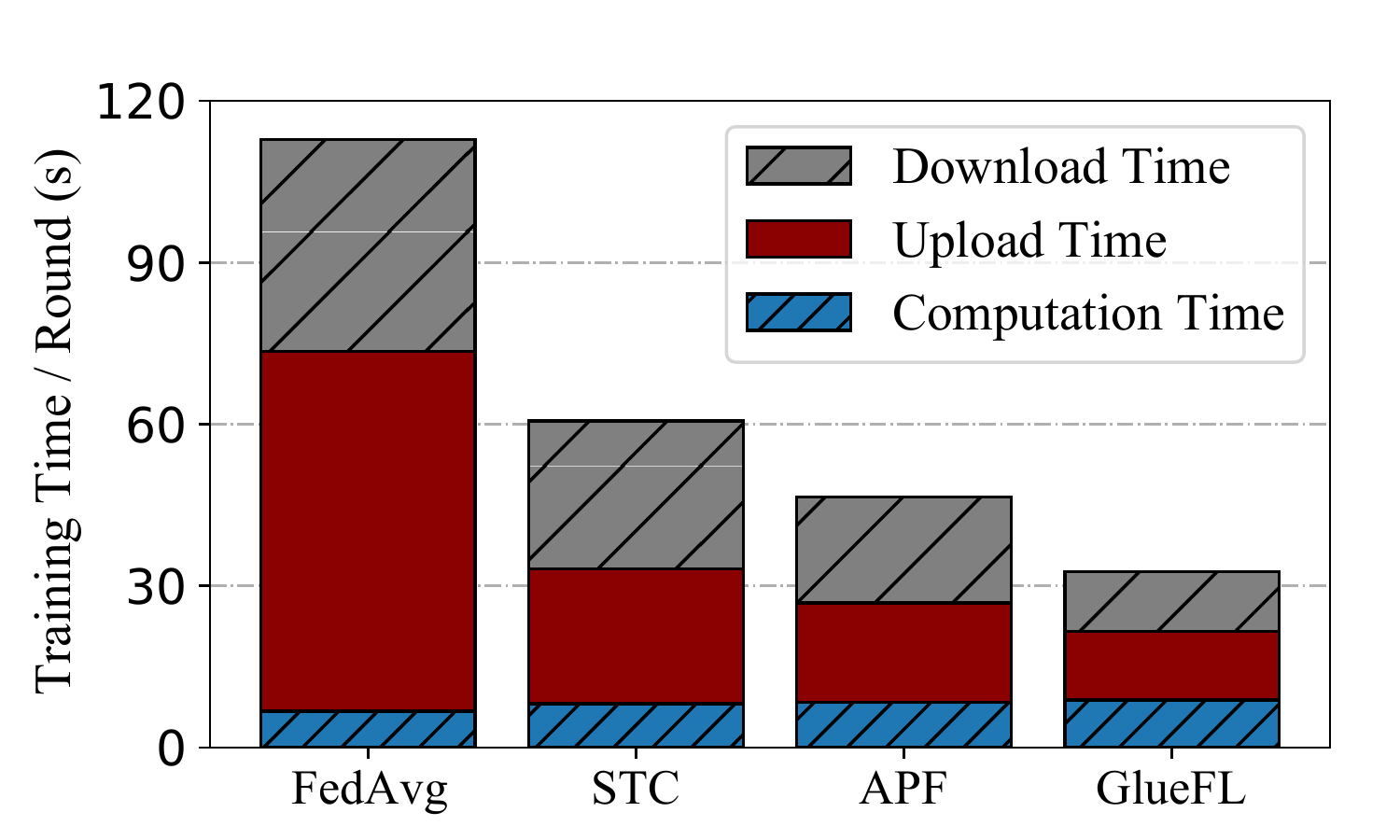}
}
\subfloat[Commercial 5G network]{
\centering
\includegraphics[width=0.29\textwidth]{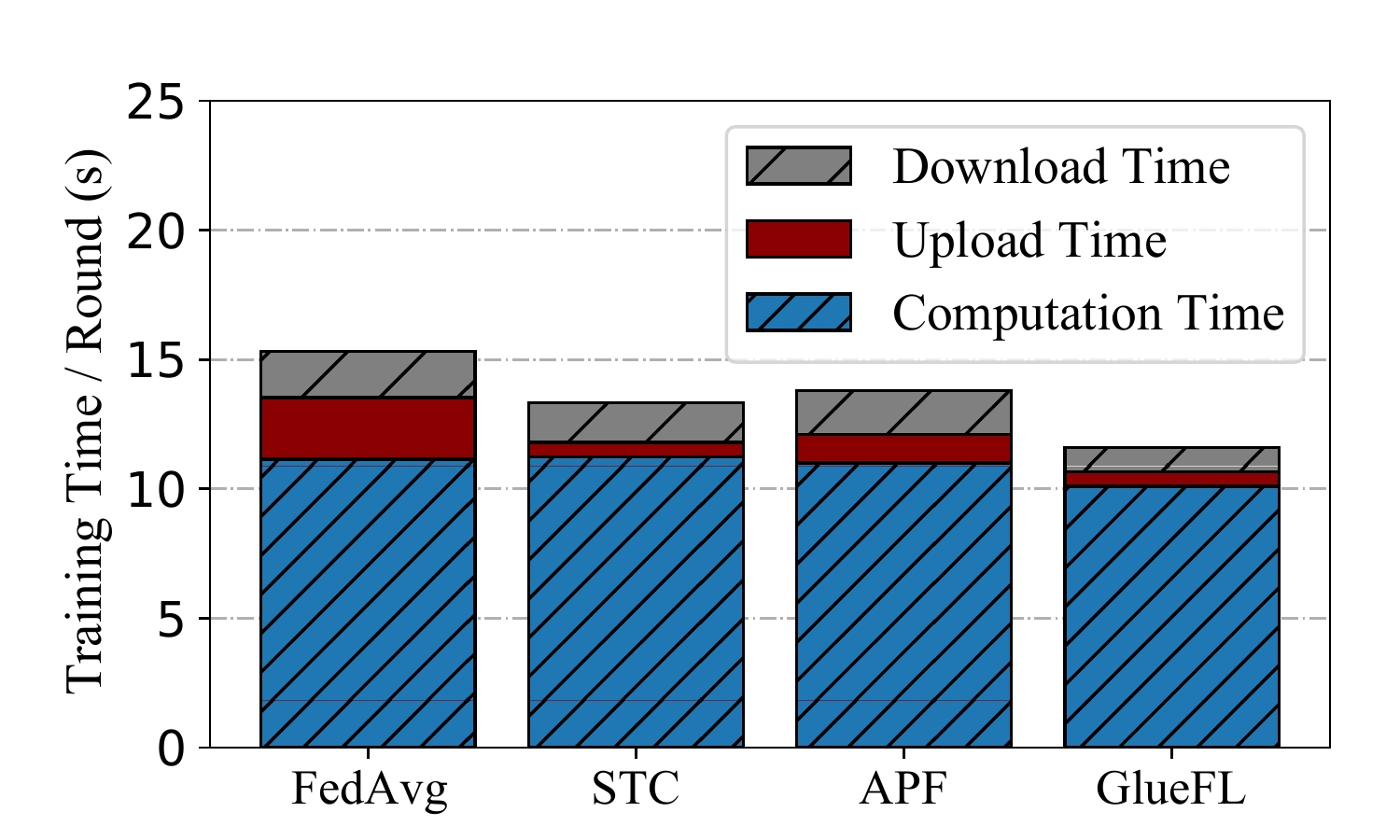}
}
\subfloat[Google Cloud datacenter network]{
\centering
\includegraphics[width=0.29\textwidth]{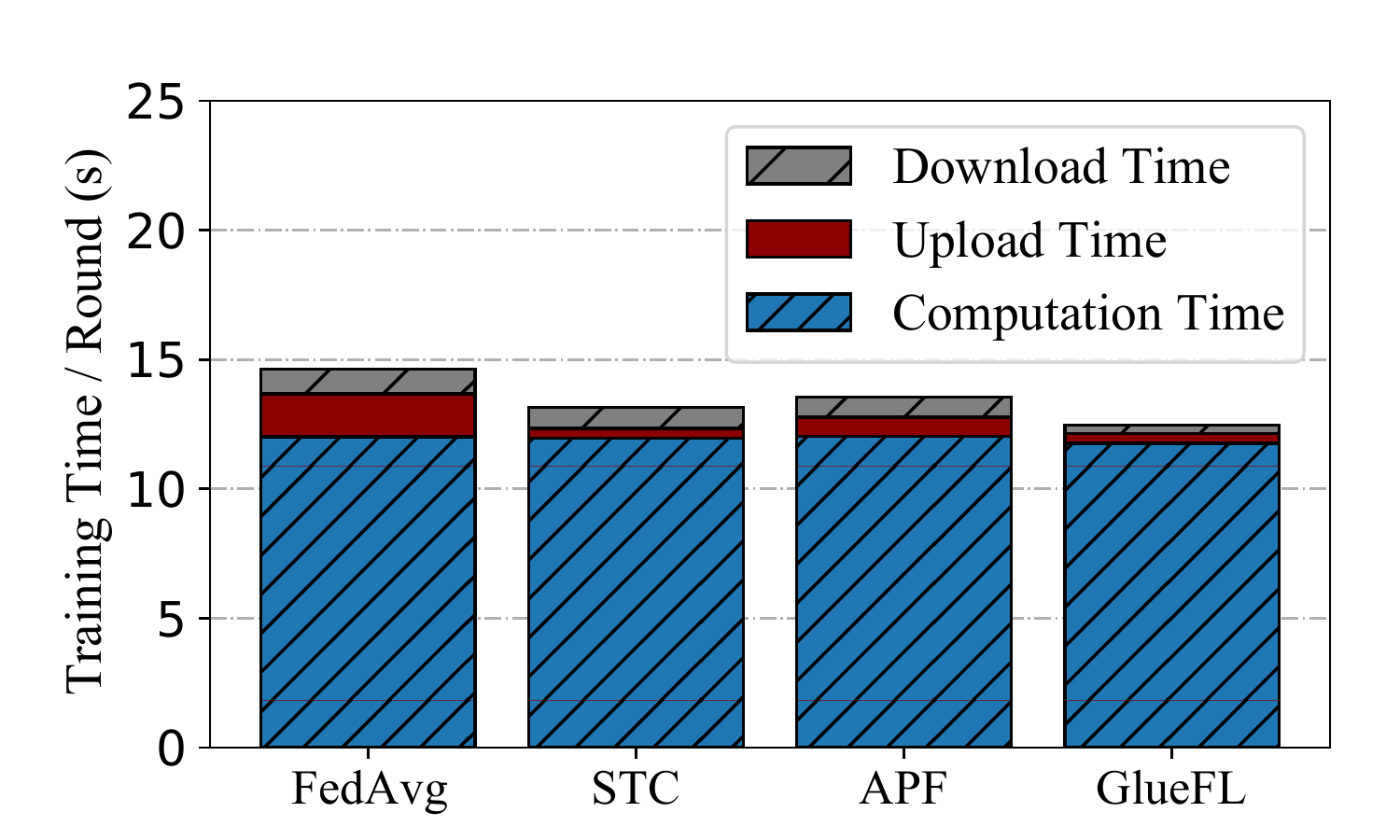}
}
\vspace{-3px}
\caption{Average share of time spent per round downloading (grey), uploading (red) and computing (blue). }

\label{fig:round_time}
\vspace{-10pt}
\end{figure*}

\subsection{Network Environment}


To further test our framework on high-throughput environments, we repeated the experiment in \Cref{table:accuracy} on commercial 5G~\cite{narayanan2021variegated} and Google Cloud~\cite{mok2021measuring} with the default settings for \mytitle (see \S\ref{sec:experimental_setup}). ~\Cref{fig:round_time} shows the total share of download, upload, and computation time for the three environments. 

According to \Cref{fig:round_time}(a), transmission time remains a bottleneck in the end-user edge devices environment as shown in \Cref{table:accuracy}. We attribute this to low-bandwidth clients. The ratio of download to upload time increases as we introduce compression. Since clients usually download faster than upload~\cite{mlab, speedtest}: new clients in FedAvg spend 70\% more time uploading than downloading the same-sized update. However, for STC and APF, download time takes on average 8\% longer than upload, confirming the discussion in \S\ref{sec:limitations}. To address this limitation, \mytitle saves downstream bandwidth and reduces download time by at least 42\% as compared with other approaches. This is because clients in the sticky group are required to download less updates and are therefore less likely to become stragglers. 

In 5G and intra-datacenter networks, computation dominates the per-round training time. Yet, straggler clients still exist and they ultimately determine the end-to-end training time. 


\begin{figure}[!t]
\centering
\vspace{-10px}
\subfloat[ShuffleNet (FEMNIST)]{ \label{fig:apdx_reg_shf}
\centering
\includegraphics[width=0.23\textwidth]{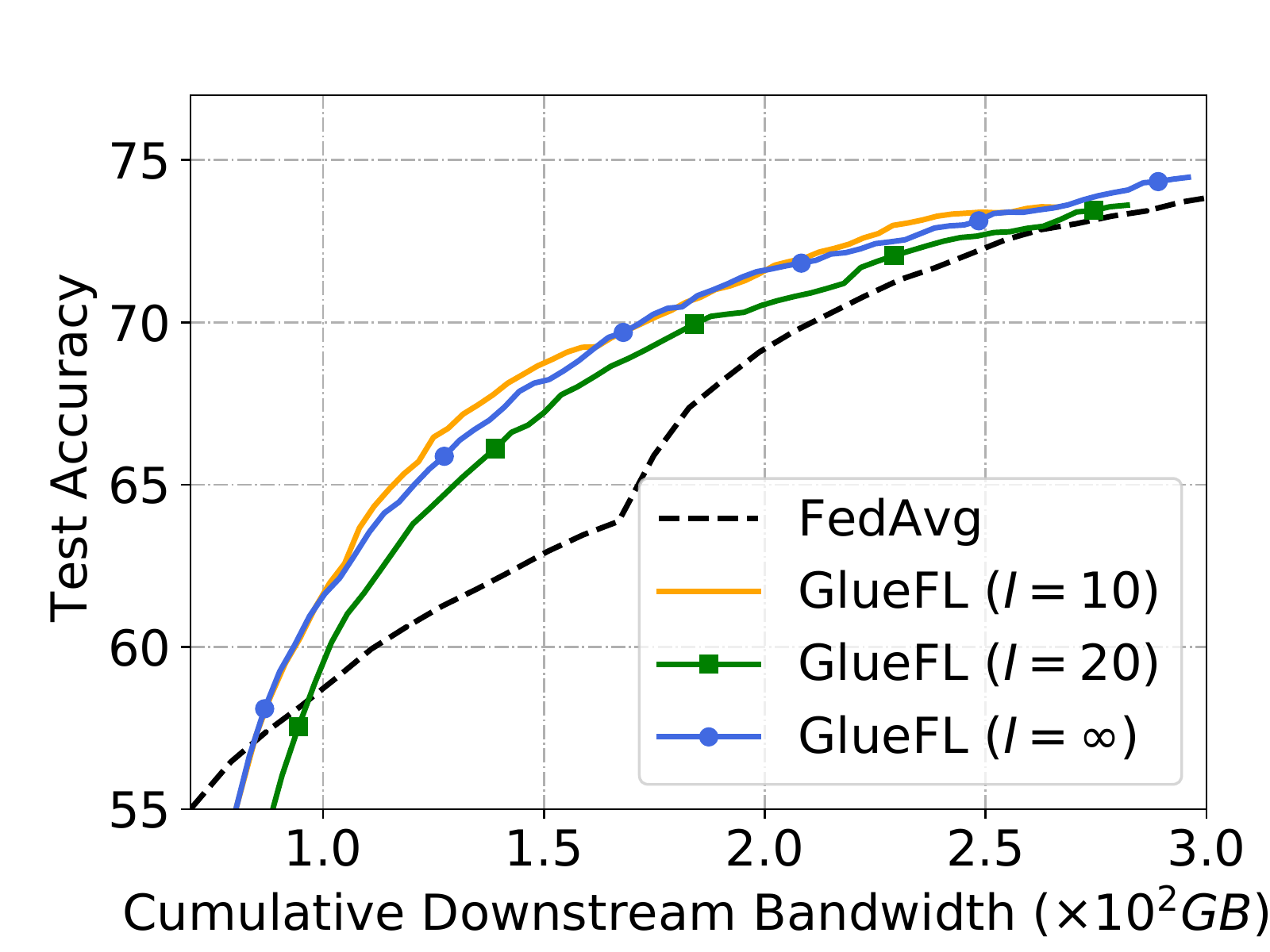}
}
\subfloat[ResNet-34 (Google Speech)]{ \label{fig:apdx_reg_res}
\centering
\includegraphics[width=0.23\textwidth]{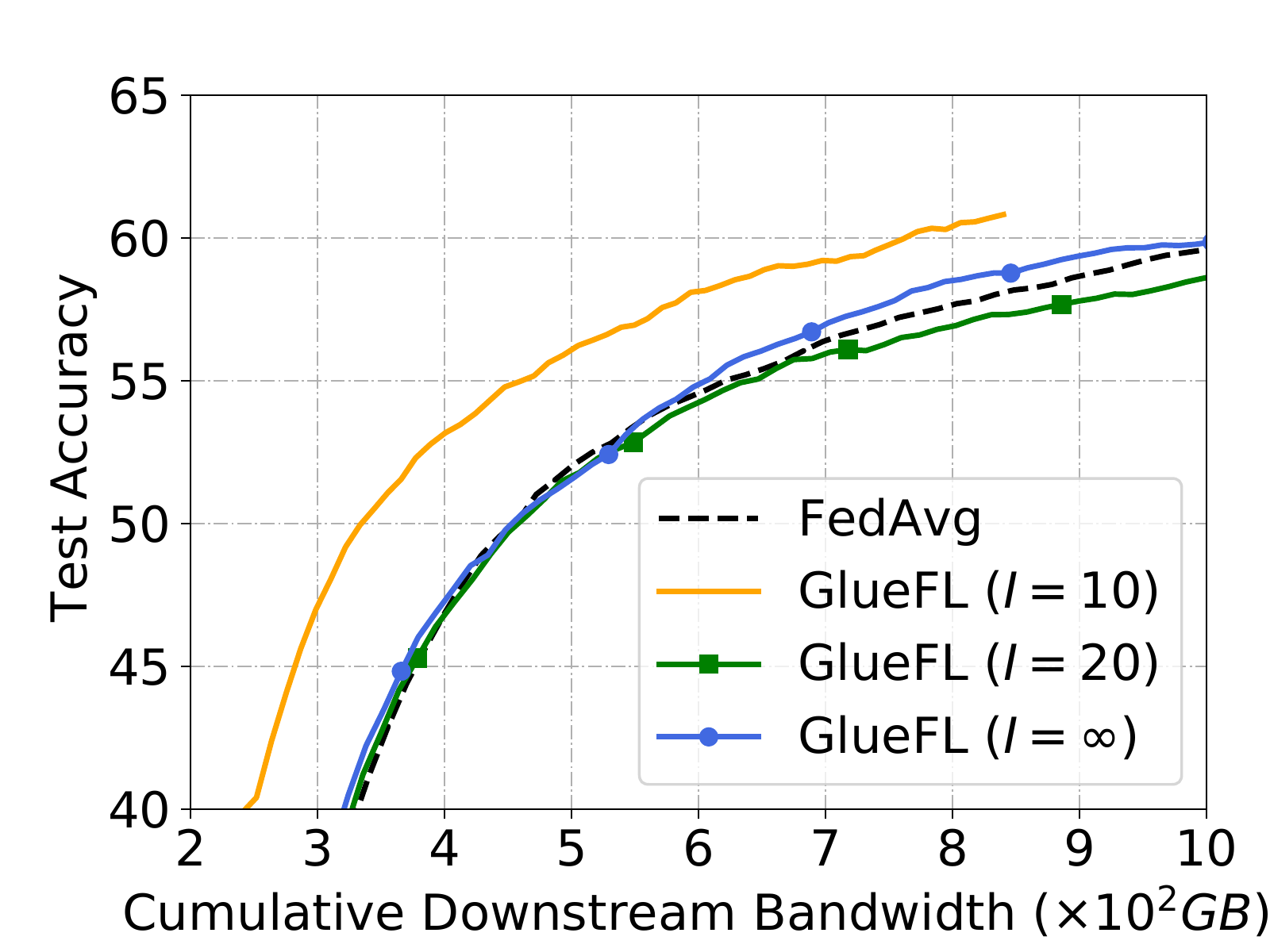}
}
\caption{Effect of shared mask regeneration. }
\end{figure}

\begin{figure}[!t]
\centering
\vspace{-10px}
\subfloat[ShuffleNet (FEMNIST)]{ \label{fig:apdx_ec_shf}
\centering
\includegraphics[width=0.23\textwidth]{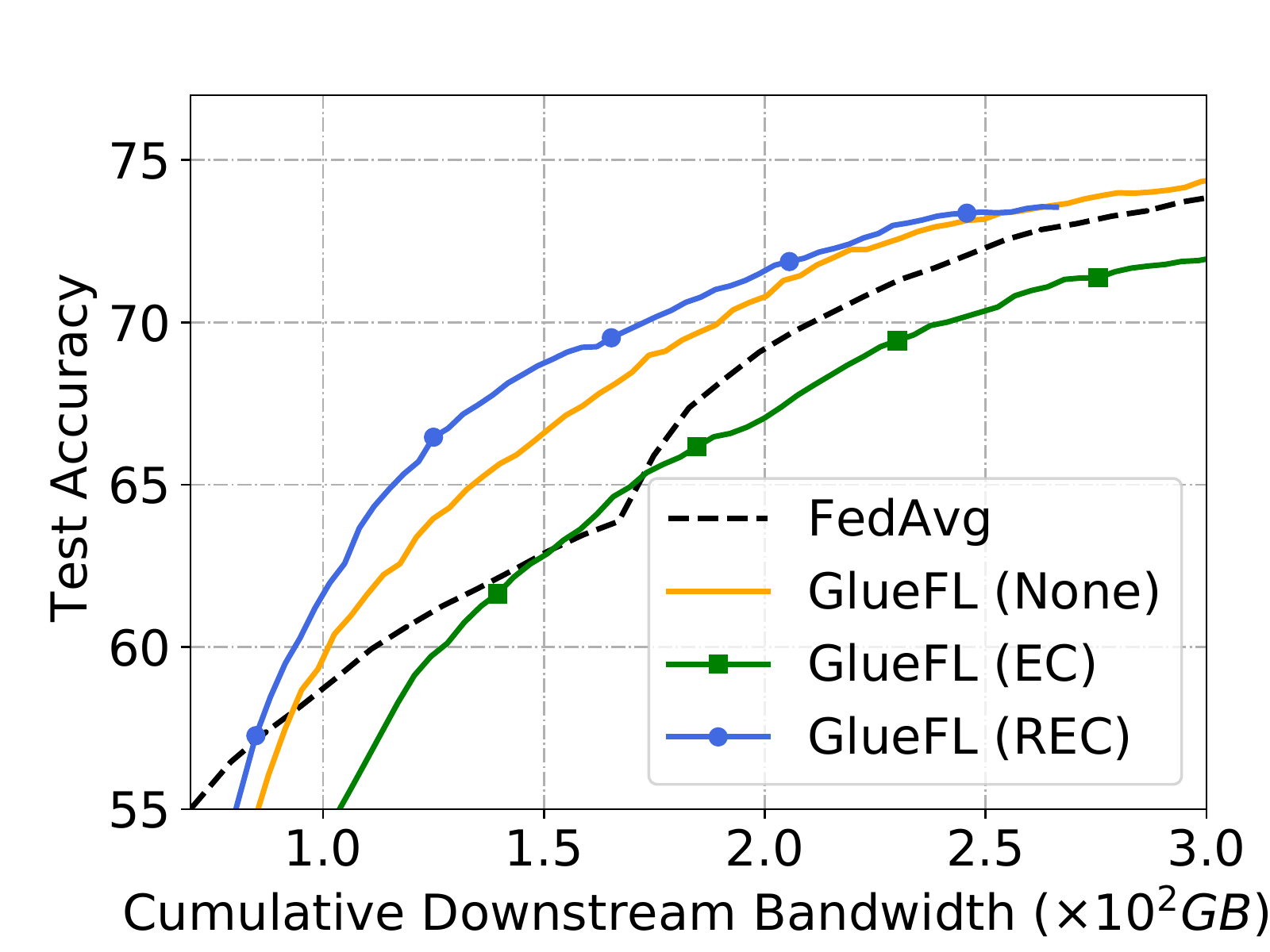}
}
\subfloat[ResNet-34 (Google Speech)]{ \label{fig:apdx_ec_res}
\centering
\includegraphics[width=0.23\textwidth]{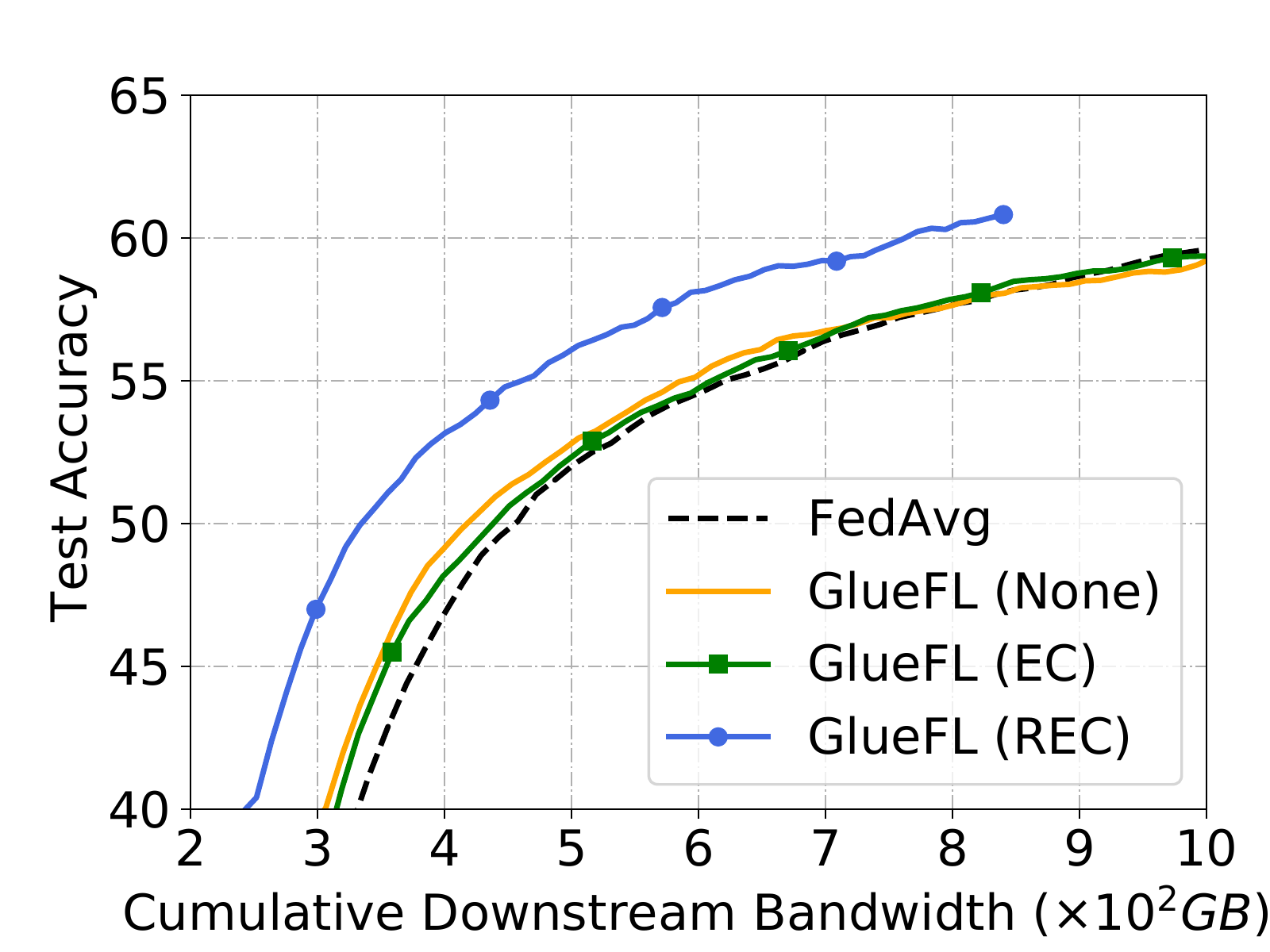}
}
\caption{Effect of error-compensation. }
\end{figure}

\subsection{Ablation Study}
We described two optimization techniques in \S\ref{sec:other-tweaks}: shared mask regeneration and error-compensation. The first technique regenerates the shared mask $M^t$ every $I$ rounds and the second technique adds a re-scaled compensation vector $h_i^{\varphi(t)}$ to local updates $\Delta_i^t$. In this section, we conduct ablation studies to evaluate the effect of these techniques.

We run \mytitle on FEMNIST with ShuffleNet and Google Speech with ResNet-34. In each round, the server samples 30 clients out of 2,800 clients (for FEMNIST) and 30 clients out of 2,066 clients (for Google Speech). For each experiment, we run 1,000 rounds and measure the downstream bandwidth and test accuracy. While \mytitle consists of both sticky sampling and mask shifting, we only change the corresponding part in mask shifting and keep other training settings the same as \S\ref{sec:experimental_setup}.

\paragraph{Shared Mask Regeneration}  As described in \S\ref{sec:other-tweaks}, we set $q_{shr}=0$ and regenerate the shared mask as $M^t \gets top_{q_{shr}}(\Tilde{\Delta}_{uni}^t)$ every $I$ rounds. A larger $I$ value indicates that $M^t$ will be regenerated less frequently. We do not regenerate $M^t$ when $I=\infty$. 

In \Cref{fig:apdx_reg_shf,fig:apdx_reg_res}, we plot results for three $I$ values: $10$, $20$, $\infty$. Both figures show that setting $I=10$ achieves the best overall performance, saving around 22\% downstream bandwidth at the target accuracy for Google Speech. The impact of $I$ on FEMNIST training in \Cref{fig:apdx_reg_shf} is less pronounced but the $I = 10$ setting still has the best accuracy. Thus, in practice, we need to set an appropriate $I$ value (e.g., $10$) to avoid a drop in accuracy. 

\paragraph{Error-Compensation} In \S\ref{sec:other-tweaks} we noted that error compensation can be used to accelerate convergence when applying compression methods in FL training. \mytitle re-scales the compensation vector $h_i^{\varphi(t)}$, following \Cref{eq:compensation}, to make it compatible with sticky sampling. In this section, we report on experiments for three error compensation settings: no compensation (\textbf{None}), compensation without re-scaling (\textbf{EC}), compensation with re-scaling (\textbf{REC}). The convergence results are shown in \Cref{fig:apdx_ec_shf,fig:apdx_ec_res}. Both figures show that removing re-scaling from error compensation immediately breaks \mytitle and harms the convergence performance. This demonstrates that it is necessary to apply re-scaling with error compensation.

\subsection{Availability and Stragglers}
In \S\ref{sec:experimental_setup} we discussed a default value of 1.3 for \texttt{over-commitment}. This means that \mytitle will sample $0.3 \times K$ additional clients to mitigate stragglers and clients that might become unavailable (e.g., go offline). In this section, we explore different values and strategies in over-commitment for GlueFL.

In \mytitle's default setting, the over-commitment applies to both sticky group $\mathcal{S}$ and non-sticky group $\mathcal{N} \setminus \mathcal{S}$. The server will sample $0.3 \times K \times (C/K)$ and $0.3 \times K \times (1-(C/K))$ additional clients from $\mathcal{S}$ and $\mathcal{N} \setminus \mathcal{S}$, respectively. However, as \mytitle only includes the fastest $K-C$ clients in all sampled non-sticky clients to $\mathcal{S}$ in each round, clients in $\mathcal{S}$ are less likely to become stragglers.
It follows that we can improve the over-commitment strategy by sampling fewer additional clients in $\mathcal{S}$ while sampling more additional clients from $\mathcal{N} \setminus \mathcal{S}$.



\begin{table}[!t] 
\centering 

\normalsize
\caption{\textbf{D}ownstream transmission \textbf{V}olume (\textbf{DV}, in $\times10^2$ GB), \textbf{D}ownload \textbf{T}ime (\textbf{DT}, in hours), \textbf{T}otal transmission \textbf{V}olume (\textbf{TV}) and \textbf{T}otal training \textbf{T}ime (\textbf{TT}) for training ShuffleNet on FEMNIST with different over-commitment (OC) settings.}
\vspace{+5pt}
\subfloat[Results of different over-commitment strategies with a constant over-commit value (1.3)]{
\label{table:oc-strategy}
\small
\begin{tabular}{cccccc}

\hline 
OC Strategy & ($S:N\setminus S$)
& \textbf{DV} & \textbf{TV} & \textbf{DT} & \textbf{TT} \\

 \hline
10\% & $1:8$ 
 & 2.1 & 3.1 & 0.6 & 2.7\\
\hline
30\% & $3:6$ 
 & 2.2 & 3.0 & 0.9 & 3.1\\
\hline
50\% & $5:4$ 
 & 2.1 & 2.9 & 1.3 & 3.8\\
\hline
$C/K$ (Default) & $7:2$ 
 & 2.2 & 3.1 & 2.2 & 5.3\\
\hline

\end{tabular}

}
\\
\normalsize
\subfloat[Results for different over-commitment values with a constant strategy (row 1 in Table~\ref{table:oc-strategy})]{
\label{table:oc-value}
\small
\begin{tabular}{ccccc}
\hline 
OC Value

& \textbf{DV} & \textbf{TV} & \textbf{DT} & \textbf{TT} \\

 \hline
 1.0 
 & 1.5 & 2.3 & 32.0 & 67.8\\
\hline
 1.1 
 & 2.2 & 3.1 & 3.5 & 10.7\\
\hline
 1.2 
 & 2.2 & 3.0 & 1.0 & 3.9\\
\hline
 1.3 
 & 2.1 & 3.1 & 0.6 & 2.7\\
\hline
 1.4 
 & 2.9 & 3.7 & 0.5 & 2.6\\
\hline
 1.5 
 & 3.1 & 4.0 & 0.5 & 2.4\\
\hline

\end{tabular}
}
\normalsize
\end{table}

\Cref{table:oc-strategy} presents the results from using four over-commitment strategies for training ShuffleNet on FEMNIST. Similar to previous tasks: we select 30 clients out of 2,800 clients in each round and we choose another 9 (i.e., $0.3 \times 30$) clients for over-commitment. We report transmission volume and training time when the model reaches the target test accuracy of 73.3\%. In the table, the OC strategy row of $10\%$ means that 1 (i.e., $0.3 \times 30 \times 10\%$), and 8 (i.e., $0.3 \times 30 \times (1-10\%)$) additional clients are sampled from $\mathcal{S}$ and $\mathcal{N} \setminus \mathcal{S}$, respectively. The results show that by choosing fewer additional clients from the sticky group, \mytitle consumes less training time without increasing the downstream bandwidth volume. 

Next, we use the best setting of 10\% (from \Cref{table:oc-strategy}) to evaluate different OC values. \Cref{table:oc-value} shows the results for OC values of 1.0 to 1.5. With increasing OC values, we find that training time decreases faster than downstream volume increases. As an example, when OC value is changed from 1.0 to 1.3, training time is decreased by 96\% and downstream volume increases by 40\%. However, increasing the OC value from 1.3 to 1.5 only reduces 11\% training time while consuming 47\% more downstream volume. In practice, one should set the OC value carefully to balance the trade-off between bandwidth and training~time. 










\section{Related Work} 
\label{sec:related_work}

The synchronization bottleneck is an established problem in FL. Existing solutions fall into roughly two categories: (1) use \emph{client sampling} to constrain the number of clients in each round; and, (2) \emph{compress model data} with strategies like sparsification and parameter freezing.
\vspace{-10px}
\paragraph{Client sampling}
FedAvg proposed a uniform sampling of clients to participate in each round. Uniform sampling has been shown to be biased, and multinomial distribution (MD) sampling was proposed to address this issue~\cite{li2020federated}. Clustered sampling~\cite{fraboni2021clustered} reduced the variance of client update aggregation by improving client representation. Oort~\cite{oort-osdi21} introduced a practical client selection algorithm, which considers both data utility and clients speed. 

\paragraph{Sparsification}
The idea of sparsification is to send only the most informative gradients. Gaia~\cite{hsieh2017gaia} transfers gradients whose absolute or relative values are larger than a given threshold. Stich et al.~\cite{stich2018sparsified}  proposed \emph{Top-K} that, given a compression ratio, selects a fraction of gradients based on their absolute values to meet the ratio. STC~\cite{sattler2019robust} extended Top-K to FL training and also uses server-side compression.

\paragraph{Parameter freezing} Parameter freezing reduces bandwidth by freezing the gradients that converged. Brock et al.~\cite{brock2017freezeout} proposed FreezeOut, which gradually froze the first few layers of a deep neural network that were observed to converge first. However, it has a coarse layer-based granularity and it degrades accuracy. APF~\cite{chen2021communication} improves on FreezeOut by freezing at a fine granularity and achieves a communication speed-up while preserving model convergence. 

Our goal with \mytitle is to coherently combine client sampling with model compression. To our knowledge, we are the first to propose a combination that is unbiased, achieves high accuracy, and lowers downstream bandwidth usage.




\section{Conclusions}
\label{sec:conclusion}
We proposed \textbf{\mytitle}, a framework to optimize downstream bandwidth in cross-device FL. \mytitle uses \emph{sticky sampling} for client selection and \emph{mask shifting} for model compression to mitigate the low download bandwidth of FL clients. We also provide a theoretical convergence guarantee for \mytitle.
In comparison with FedAvg, \mytitle achieves similar accuracy while decreasing total training time by 36\% and uses 22\% less downstream bandwidth. \mytitle also outperforms STC~\cite{sattler2019robust} and APF~\cite{chen2021communication}. 

\nocite{langley00}




\bibliography{ref}
\bibliographystyle{mlsys2023}


\newpage
\appendix
\onecolumn
\allowdisplaybreaks

\section{Analysis of Sampling Schemes} \label{proof:propsticky}

%
In this section, we provide a comparison between uniform sampling and sticky sampling to demonstrate the advantage of sticky sampling. We first analyze the probability that a client is re-sampled after $r$ rounds and then give the expected number of rounds for a client to be re-sampled. 




\subsection{Analysis on Uniform Sampling}
\begin{proposition}
\label{prop:uniform}
Suppose a client is sampled at the current round. With uniform sampling, there is a probability of $\frac{K}{N}(1-\frac{K}{N})^{r-1}$ that the client is sampled after $r$ rounds. On average, a client is sampled every $N/K$ rounds. 
\begin{proof}
    The client is sampled with a probability of $\frac{K}{N}$. The client has not been selected for the first $(r-1)$ rounds. Thus, this happens with a probability of $\frac{K}{N}(1-\frac{K}{N})^{r-1}$. Furthermore, the value of averaged sampled rounds is $\sum_{r=1}^{\infty} \frac{K}{N}(1-\frac{K}{N})^{r-1} \cdot r = N/K$. 
\end{proof}
\end{proposition}


\subsection{Analysis on Sticky Sampling}
\begin{proposition} \label{prop:sticky}
Suppose a client is sampled at the current round. Using sticky sampling, the client in the sticky group is sampled with a probability of $\frac{1}{(N-S)K - (K-C)S} (\frac{K(NC-SK)}{S} (1 - \frac{K}{S})^{r-1} + (K-C)^2 (1 - \frac{K-C}{N-S})^{r-1})$ after $r$ rounds. As expected, the client trains a model every $N/K$ rounds. 
\begin{proof}
    In the sticky group, a client is sampled or moved to the non-sticky group with the probability of $\frac{C}{S}$ and $\frac{K-C}{S}$, respectively. And, a client is sampled from the non-sticky group with probability $\frac{K-C}{N-S}$.

    There are two strategies to sample a client that has participated in model training. First, it is sampled from the sticky group, where the probability is $\frac{C}{S}(\frac{S-K}{S})^{r-1}$ after $r$ rounds. Second, it is sampled from the non-sticky group, indicating the client is moved out of the sticky group in the middle. Therefore, the probability is $\sum_{i=1}^{r-1} (1-\frac{K-C}{N-S})^{i-1} \cdot \frac{K-C}{N-S} \cdot (\frac{S-K}{S})^{r-i-1} \cdot (\frac{K-C}{S}) = \frac{(K-C)^2}{(N-S)K - (K-C)S} ((1-\frac{K-C}{N-S})^{r-1} - (\frac{S-K}{S})^{r-1})$. By summing up these two probabilities, we can obtain the desired result. Furthermore, similar to \Cref{prop:uniform}, we can calculate the value of averaged sampled rounds. 
\end{proof}
\end{proposition}

\subsection{Discussion}
According to the proof of \Cref{prop:sticky}, the probability of a client in the sticky group being sampled after $r$ rounds is greater or equal to $\frac{C}{S}(\frac{S-K}{S})^{r-1}$, which is the probability that it is still sampled from the sticky group.
Then, for 
$r \in \{1, \dots, 1 + \left\lfloor \left(\log{\frac{CN}{SK}}\right) / \left(\log\frac{S(N-K)}{N(S-K)}\right) \right\rfloor\}$,
$\frac{C}{S}(\frac{S-K}{S})^{r-1}$ is greater or equal to $\frac{K}{N}(1-\frac{K}{N})^{r-1}$, the probability that a client is sampled after $r$ rounds in uniform sampling (\Cref{prop:uniform}).

\section{Some Useful Lemmas} 
In this section, we provide two useful lemmas, which will apply to our subsequent analysis in \S\ref{sec:proof_of_convergence}. \Cref{lemma:lipschitz} is used to present the progress in one single step in FL (\S\ref{sec:progress_single_step}) and \Cref{lemma:min_exp} is used to bound the gap between two successive global models (\Cref{lemma:glueFL_no_mask_global}). 
\begin{lemma}[\cite{bottou2018optimization}] \label{lemma:lipschitz}
Suppose a function $\mathcal{H}$ is $L_c$-continuous and $L_s$-smooth. For any $w, v \in \mathbb{R}^d$, the following inequality holds for $\mathcal{H}$: 
\begin{align*}
    \|\nabla \mathcal{H}(w)\|_2 \leq L_c; \qquad \mathcal{H}(w) \leq \mathcal{H}(v) + \left\langle\nabla \mathcal{H}(v), w - v\right\rangle + \frac{L_s}{2} \|w-v\|_2^2
\end{align*}
\end{lemma}

\begin{lemma}[Lemma 4 in \cite{karimireddy2020scaffold}] \label{lemma:split_array} \label{lemma:min_exp}
Let $\varepsilon = \{\varepsilon_1, \dots, \varepsilon_a\}$ be $a$ random variables in $\mathbb{R}^d$, which are not assumed to be independent. If $\E\left[\varepsilon_i\right]=e_i$, and the variance is bounded by $\E\left[\norm{\varepsilon_i-e_i}\right]\leq\sigma^2$, we have:
\begin{equation*}
\E\left[\norm{\sum_{i=1}^a\varepsilon_i}\right]\leq\norm{\sum_{i=1}^a e_i}+a^2\sigma^2
\end{equation*}
If we further suppose that $\E\left[\varepsilon_i|\varepsilon_{i-1},\dots,\varepsilon_{1}\right]=e_i$, in which case the $\{\varepsilon_i-e_i\}$ form a martingale difference sequence, and the bound of the variance $\E\left[\norm{\varepsilon_i-e_i}\right]\leq\sigma^2$ holds, we have the following, tighter bound:
\begin{equation*}
\E\left[\norm{\sum_{i=1}^a\varepsilon_i}\right]\leq2\norm{\sum_{i=1}^a e_i}+2a\sigma^2
\end{equation*}


\end{lemma}

\section{Proof of \Cref{theorem:convergence}}
\label{sec:proof_of_convergence}
In this section, we theoretically analyze the convergence rate of sticky-sampling in \mytitle on non-convex functions, under \Cref{ass:L-smooth,ass:bounded_var}. The conclusion has been mentioned in \Cref{theorem:convergence}.
The proof follows the same template as those of \cite{bottou2018optimization, karimireddy2020scaffold, wang2020tackling, yang2021achieving}, and proceeds as follows: (1) we use \Cref{lemma:lipschitz} to bound the expected progress in each step (\S\ref{sec:progress_single_step}) by a sum of two terms.
(2) We bound the first term through a bound on local updates (\Cref{sec:local-update}) and our unbiased aggregation.
(3) We bound the second term by adapting a bound on the norm between two consecutive models to account for our aggregation weights (\S\ref{sec:norm-update-bound}).
(4) We use the bound on the expected progress in each step in a telescopic sum to bound the overall progress over training (\S\ref{sec:convergence-proof}).

We first present steps (1) and (4) in \Cref{sec:progress_single_step,sec:convergence-proof}, which represent the high level articulation of the proof, before presenting the lower level results for steps (2) and (3) in \Cref{sec:local-update,sec:norm-update-bound}.

\newcommand{\selsticky}[1]{\mathcal{C}^{#1}}
\newcommand{\sticky}[1]{\mathcal{S}^{#1}}
\newcommand{\selnonsticky}[1]{\mathcal{R}^{#1}}
\newcommand{\nonsticky}[1]{\mathcal{N} \setminus \mathcal{S}^{#1}}
\newcommand{\innerp}[2]{\left\langle#1, #2\right\rangle}

\subsection{Progress in one single step}
\label{sec:progress_single_step}

We first bound the expected progress after one step of the model update. By definition, $\w^{t+1} = \w^t - \gamma \sum_{i \in \mathcal{K}^t} \nu_{i}^t \sum_{e=0}^{E-1} \cdot \g_i^{t, e}$, where $\nu_{i}^t$ can be either $\nu_{i,s}^t$ or $\nu_{i,r}^t$ depending on the client's membership. Since all local objective functions are $L_s$-smooth, the global objective $F$ is $L_s$-smooth as well. Thus, according to Lemma \ref{lemma:lipschitz}, we have:
\begin{align} \label{eq:full}
    \conE\Big[F\left(\w^{t+1}\right)\Big] - F(\w^t) \leq \underbrace{\conE \left\langle \nabla F(\w^t), \w^{t+1} - \w^t \right\rangle}_{\q_1} + \frac{L_s}{2} \underbrace{\conE \left\| \w^{t+1} - \w^t \right\|_2^2}_{\q_2} 
\end{align}
 where $\conE$ means the expected value at round $(t+1)$, condition on all information at round $t$, including the model $\w^t$ and the participants $\mathcal{K}^{t-1}$. The expectation is over the randomness of client selection ($\mathcal{K}^{t}$) and batch selection at the client's ($\xi_i \sim \mathcal{D}_i$ from \Cref{sec:background_fl}).

We first provide the upper bound analysis for term $\mathcal{Q}_1$. Intuitively, our unbiased aggregation combines with a technical client local drift bound adapted from previous work (\Cref{sec:local-update}) to decompose this term.
Remember that as \Cref{theorem:unbiased} indicates, our weighted update is an unbiased estimate of the true update over all clients. That is:
\begin{align}
\conE\Big[ \w^{t+1} - \w^t \Big]
= \conE\Big[ \Delta^{t} \Big]
= \sum_{i=1}^N p_i \E_{\xi_i \sim \mathcal{D}_i}\big[\Delta^{t}_{i}\big] ,
\end{align}

where we decomposed $\conE$ in the randomness over client sampling, and local updates. The  expectation in the right-hand side is over the local training steps of each client. Based on the form of local updates,
we have that $\E_{\xi_i \sim \mathcal{D}_i}\big[\w_i^{t, E} - \w^{t}\big] = - \gamma \sum_{e=0}^{E-1} \E_{\xi_i \sim \mathcal{D}_i}\big[ \g_i^{t,e} \big]$. Considering the unbiased estimation assumption mentioned in Section \ref{sec:background_fl}, we have that $\forall e, i: \ \E_{\xi_i \sim \mathcal{D}_i}\big[\g_i^{t, e}\big] = - \nabla F_i (\w_i^{t, e})]$.
Therefore, the term $\mathcal{Q}_1$ above can be bounded as follows: 
\begin{align}
    \mathcal{Q}_1 &= \conE \left\langle \nabla F(\w^t), \w^{t+1} - \w^t \right\rangle \\
    &= \left\langle \nabla F(\w^t), - \gamma \sum_{i=1}^N p_i \cdot \left( \sum_{e=0}^{E-1} \E_{\xi_i \sim \mathcal{D}_i}\big[ g^{t,e}_i \big] \right) \right\rangle \\
    &= -\gamma E \cdot \left\langle \sum_{i=1}^N p_i \nabla F_i(\w^t),  \sum_{i=1}^N \sum_{e=0}^{E-1} \frac{p_i}{E} \E_{\xi_i \sim \mathcal{D}_i}\big[ \nabla f_i\left(\w^{t,e}_i\right) \big] \right\rangle \\
    &= -\frac{\gamma E}{2} \cdot \norm{\nabla F(\w^t)} - \frac{\gamma E}{2} \norm{\sum_{i=1}^N \sum_{e=0}^{E-1} \frac{p_i}{E} \nabla F_i\left(\w^{t,e}_i\right)} + \frac{\gamma E}{2} \norm{\sum_{i=1}^N \sum_{e=0}^{E-1} \frac{p_i}{E} \E_{\xi_i \sim \mathcal{D}_i}\big[ \nabla f_i(\w^t) - \nabla f_i\left(\w^{t,e}_i\right)\big]} \\
    &\leq -\frac{\gamma E}{2} \cdot \norm{\nabla F(\w^t)}  - \frac{\gamma}{2 E} \norm{\sum_{i=1}^N \sum_{e=0}^{E-1} p_i \nabla F_i\left(\w^{t,e}_i\right)} + \frac{\gamma E}{2} \cdot \sum_{i=1}^N \sum_{e=0}^{E-1} \frac{p_i}{E} \E_{\xi_i \sim \mathcal{D}_i}\big[ \norm{\nabla f_i(\w^t) - \nabla f_i\left(\w^{t,e}_i\right)} \big] \\
    &\leq -\frac{\gamma E}{2} \cdot \norm{\nabla F(\w^t)} -\frac{\gamma}{2 E} \norm{\sum_{i=1}^N \sum_{e=0}^{E-1} p_i \nabla F_i\left(\w^{t,e}_i\right)} + \frac{\gamma L_s^2}{2} \cdot \sum_{i=1}^N \sum_{e=0}^{E-1} p_i \E_{\xi_i \sim \mathcal{D}_i}\big[ \norm{\w^t - \w^{t,e}_i} \big]
\end{align}
where the last equality follows from the fact that $\langle a, b \rangle = \frac{1}{2} a^2 + \frac{1}{2} b^2 - \frac{1}{2} (a - b)^2$ and the assumption we make in \Cref{subsec:assumption} that $\E_{\xi_i \sim \mathcal{D}_i}\big[ \nabla f_i(\mathbf{w}, \xi_i) \big] = \nabla F_i(\mathbf{w})$; the first inequality follows from Jensen's Inequality because $\sum_{i=1}^N \sum_{e=0}^{E-1} \frac{p_i}{E} = 1$; and the next inequality from the $L_s$-smoothness assumption. 

Plugging \Cref{lemma:glueFL_no_mask_local} into the above bound for $\mathcal{Q}_1$, and using \Cref{lemma:glueFL_no_mask_global} to bound $\mathcal{Q}_2$, we have that: 
\begin{align}
    &\conE \left(F\left(\w^{t+1}\right)\right) - F\left(\w^t\right) \\
    \leq & -\frac{\gamma E}{2} \|\nabla F(\w^t)\|_2^2 + \frac{3 \gamma^3 E^2 L_s^2}{2} (E L_c + \sigma^2) + \frac{L_s \gamma^2}{2} \cdot \mathbb{E} \left( \sum_{i \in \mathcal{C}^t} \frac{S^2}{C^2} p_i^2 + \sum_{i \in \mathcal{R}^t} \left(\frac{N-S}{K-C}\right)^2 p_i^2\right) E \sigma^2 \\
    & + \frac{L_s \gamma^2}{2} E^2 L_c^2 \left(\sum_{i \in \sticky{t}} \frac{S}{C}p_i^2 + \sum_{i \in \nonsticky{t}} \frac{N-S}{K-C}p_i^2 \right) - \left(\frac{\gamma}{2 E} - \frac{L_s \gamma^2}{2}\right) \norm{\sum_{i=1}^N \sum_{e=0}^{E-1} p_i \nabla F_i\left(\w^{t,e}_i\right)}
\end{align}

\subsection{Final Convergence Result}
\label{sec:convergence-proof}

Let $\gamma \leq \frac{1}{E L_s}$. By averaging the above inequality over $t$ from $1$ to $T$, we have:
\begin{align}
    &\quad \frac{1}{T} \sum_{t=1}^T \conE (F(\w^{t+1}) - F(\w^{t})) \\
    &\leq  -\frac{\gamma E}{2 T} \|\nabla F(\w^t)\|_2^2 + \frac{3 \gamma^3 E^2 L_s^2}{2} (E L_c + \sigma^2) + \frac{L_s \gamma^2 E \sigma^2}{2T} \sum_{t=1}^T \mathbb{E} \left( \sum_{i \in \mathcal{C}^t} \frac{S^2}{C^2} p_i^2 + \sum_{i \in \mathcal{R}^t} \left(\frac{N-S}{K-C}\right)^2 p_i^2\right) \\
    &\quad + \frac{L_s \gamma^2 E^2 L_c^2}{2 T} \sum_{t=1}^T \left(\sum_{i \in \sticky{t}} \frac{S}{C}p_i^2 + \sum_{i \in \nonsticky{t}} \frac{N-S}{K-C}p_i^2 \right) \\
    &= -\frac{\gamma E}{2 T} \|\nabla F(\w^t)\|_2^2 + \frac{3 \gamma^3 E^2 L_s^2}{2} (E L_c + \sigma^2) + \frac{L_s \gamma^2 E (\sigma^2 + E L_c^2)}{2T} \sum_{t=1}^T \left( \sum_{i = 1}^N \frac{S^2}{C N} p_i^2 + \sum_{i = 1}^N \frac{(N-S)^2}{N(K-C)} p_i^2\right) 
\end{align}

where the last equation follows that (i) a client in the sticky group and the non-sticky group with the probability of $\frac{S}{N}$ and $\frac{N-S}{N}$, respectively; (ii) a client is sampled from the sticky group and the non-sticky group with the probability of $\frac{C}{S}$ and $\frac{K-C}{N-S}$, respectively. Therefore, the convergence rate is 
\begin{align}
    \frac{1}{T} \sum_{t=1}^T \|\nabla F(\w^t)\|_2^2 &\leq \frac{2(F(\w^1) - F_*)}{\gamma E T} + 3 \gamma^2 E L_s^2 (E L_c + \sigma^2) + \frac{L_s  \gamma (\sigma^2 + E L_c^2)}{N} \left( \frac{S^2}{C} + \frac{(N-S)^2}{K-C} \right) \sum_{i=1}^N p_i^2 
\end{align}
By setting the learning rate as devised in Theorem \ref{theorem:convergence}, we can obtain the desired result. 

\subsection{Bounded Gap between two successive local updates.}
\label{sec:local-update}

\begin{lemma} \label{lemma:glueFL_no_mask_local}
Suppose, for all $i \in \{1, \dots, N\}$, the local objective function $F_i$ is $L_c$-continuous and $L_s$-smooth. Then, for all $e \in \{0, \dots, E-1\}$, we have 
\begin{align}
    \mathbb{E} \| \w_i^{t,e} - \w^t \|_2^2 \leq 3E \left(E \gamma^2 L_c^2 + \gamma^2 \sigma^2\right)
\end{align}
\begin{proof}
As we know, the recurrence formula for $\w^{t, e}_i = \w^{t, e-1}_i - \gamma g_i^{t,e-1}$. Through this relationship, we can bound for $\E\left\|\w^{t,e}_i - \w^t\right\|_2^2$,
\begin{align}
    \mathbb{E} \| \w_i^{t,e} - \w^t \|_2^2 &= \mathbb{E} \| \w_i^{t,e-1} - \gamma g_i^{t,e-1} - \w^t \|_2^2 \\
    & \overset{(a)}{=} \mathbb{E} \| \w_i^{t,e-1} - \w^t - \gamma \nabla F_i (\w_i^{t, e-1}) \|_2^2 + \gamma^2 \cdot \E\left\| g_i^{t,e-1} - \nabla F_i (\w_i^{t, e-1})\right\|_2^2 \\ \label{eq:14}
    & \overset{(b)}{\leq} \left(1 + \frac{1}{E-1}\right) \cdot \mathbb{E} \| \w_i^{t,e-1} - \w^t \|_2^2 + E \gamma^2 \cdot \E \| \nabla F_i(\w_i^{t,e-1}) \|_2^2 + \gamma^2 \cdot \E \left\| g_i^{t,e-1} - \nabla F_i (\w_i^{t, e-1})\right\|_2^2 \\
    & \overset{(c)}{\leq} \left(1 + \frac{1}{E-1}\right) \cdot \mathbb{E} \| \w_i^{t,e-1} - \w^t \|_2^2 + E \gamma^2 L_c^2 + \gamma^2 \sigma^2 \\
    & \leq \sum_{\varphi=0}^{e-1} \left(1 + \frac{1}{E-1}\right)^\varphi \cdot(E \gamma^2 L_c^2 + \gamma^2 \sigma^2) \\ 
    & \leq 3E \left(E \gamma^2 L_c^2 + \gamma^2 \sigma^2\right)
\end{align}
In the above proof, equation ($a$) separates the mean and the variance, the first inequality ($b$) uses $(a+b)^2 \leq (1+\alpha)a^2 + (1+\frac{1}{\alpha})b^2$, and the inequality ($c$) follows \Cref{ass:bounded_var,ass:L-smooth}. 
\end{proof}
\end{lemma}



\subsection{Bounded gap between two successive global models}
\label{sec:norm-update-bound}

Inspired by the proof of Theorem 2 in \cite{yang2021achieving}, we derive the following lemma to bound $Q_2$ accounting for \mytitle reweighted aggregation in \Cref{algo:sticky_sampling}: 


\begin{lemma} \label{lemma:glueFL_no_mask_global}
Suppose \Cref{ass:bounded_var} and \ref{ass:L-smooth} hold. With \Cref{algo:sticky_sampling} by setting the weights $\nu^t_{i,s} = \frac{S}{C} p_i$ and $\nu^t_{i,r} = \frac{N-S}{K-C} p_i$ mentioned in Section \ref{sec:sticky_sampling}, let $\alpha_i = p_i \sum_{e=0}^{E-1} \nabla F_i (w_i^{t, e})$, the bound for two successive models should be 
\begin{align}
    \conE \left\| \w^{t+1} - \w^t \right\|_2^2 \leq & \gamma^2 E \sigma^2 \conE \left(\sum_{i \in \mathcal{C}^t} \left(\frac{S}{C} p_i\right)^2 + \sum_{i \in \mathcal{R}^t} \left(\frac{N-S}{K-C} p_i\right)^2 \right) \nonumber \\ 
    &+ \gamma^2 \conE \left(\frac{S}{C} \sum_{i \in \sticky{t}} p_i^2 E^2 L_c^2 + \frac{N-S}{K-C} \sum_{i \in \nonsticky{t}} p_i^2 E^2 L_c^2 + \norm{\sum_{i=1}^N \alpha_i} \right)
\end{align}

\begin{proof}
As we know, the relationship between two successive models is 
\begin{align}
    &\quad \conE \left\| \w^{t+1} - \w^t \right\|_2^2 \\
    &= \gamma^2 \cdot \conE \left \| \sum_{i \in \mathcal{C}^t} \nu^t_{i,s} \sum_{e=0}^{E-1} g_i^{t,e} + \sum_{i \in \mathcal{R}^t} \nu^t_{i,r} \sum_{e=0}^{E-1} g_i^{t,e} \right \|_2^2 \\
    &\leq \gamma^2 \conE \left(\sum_{i \in \mathcal{C}^t} (\nu_{i, s}^t)^2 + \sum_{i \in \mathcal{R}^t} (\nu_{i, r}^t)^2 \right) \cdot E \sigma^2 + \gamma^2 \conE \norm{\sum_{i \in \mathcal{C}^t} \nu^t_{i,s} \sum_{e=0}^{E-1} \nabla F_i(\w_i^{t,e}) + \sum_{i \in \mathcal{R}^t} \nu^t_{i,r} \sum_{e=0}^{E-1} \nabla F_i(\w_i^{t,e})} \label{eq:00014}
\end{align}
where the inequality is based on Lemma \ref{lemma:min_exp}. Next, we ignore the coefficient and find the bound for the second term of \Cref{eq:00014} by plain expanding the term as proposed in \cite{yang2021achieving}:
Let $\alpha_i = p_i \sum_{e=0}^{E-1} \nabla F_i (w_i^{t, e})$, and since $\nu^t_{i,s} = \frac{S}{C} p_i$ and $\nu^t_{i,r} = \frac{N-S}{K-C} p_i$, we have
\begin{align}
    &\quad \conE \norm{\sum_{i \in \mathcal{C}^t} \nu^t_{i,s} \sum_{e=0}^{E-1} \nabla F_i(\w_i^{t,e}) + \sum_{i \in \mathcal{R}^t} \nu^t_{i,r} \sum_{e=0}^{E-1} \nabla F_i(\w_i^{t,e})} = \conE \norm{\sum_{i \in \mathcal{C}^t} \frac{S}{C} \alpha_i + \sum_{i \in \mathcal{R}^t} \frac{N-S}{K-C} \alpha_i} \\
    &= \conE \left(\underbrace{\sum_{i \in \mathcal{C}^t} \norm{\frac{S}{C} \alpha_i}}_{\text{$C$ terms}} + \underbrace{\sum_{i \in \mathcal{R}^t} \norm{\frac{N-S}{K-C} \alpha_i}}_{\text{$(K-C)$  terms}} + \underbrace{\sum_{i \neq j, i, j \in \selsticky{t}} \left(\frac{S}{C}\right)^2 \innerp{\alpha_i}{\alpha_j}}_{\text{$C(C-1)$ terms}} + \underbrace{\sum_{i \neq j, i, j \in \selnonsticky{t}} \left(\frac{N-S}{K-C}\right)^2 \innerp{\alpha_i}{\alpha_j}}_{\text{$(K-C)(K-C-1)$ terms}}  \right. \nonumber \\
    &\qquad \qquad \quad \left. + 2 \underbrace{\sum_{i \in \selsticky{t}, j \in \selnonsticky{t}} \left(\frac{S}{C}\right) \left(\frac{N-S}{K-C}\right) \innerp{\alpha_i}{\alpha_j}}_{\text{$C(K-C)$ terms}} \right) \label{lemma_eq:22}
\end{align}

Before analyzing the bound of \Cref{lemma_eq:22}, we provide the constant results for the following expectations: 
\begin{align}
    \E \norm{\alpha_i} &= \frac{1}{S} \sum_{s \in \sticky{t}} \norm{\alpha_s}, \text{for $i \in \sticky{t}$} \\
    \E \norm{\alpha_i} &= \frac{1}{N-S} \sum_{r \in \nonsticky{t}} \norm{\alpha_r}, \text{for $i \in \nonsticky{t}$} \\
    \E \innerp{\alpha_i}{\alpha_j} &= \frac{1}{S^2} \sum_{s_1, s_2 \in \sticky{t}} \innerp{\alpha_{s_1}}{\alpha_{s_2}}, \text{for $i, j \in \sticky{t}$}\\
    \E \innerp{\alpha_i}{\alpha_j} &= \frac{1}{(N-S)^2} \sum_{r_1, r_2 \in \nonsticky{t}} \innerp{\alpha_{r_1}}{\alpha_{r_2}}, \text{for $i, j \in \nonsticky{t}$} \\
    \E \innerp{\alpha_i}{\alpha_j} &= \frac{1}{S(N-S)} \sum_{s \in \sticky{t}, r \in \nonsticky{t}} \innerp{\alpha_{s}}{\alpha_{r}}, \text{for $i \in \sticky{t}, j \in \nonsticky{t}$} 
\end{align}
Therefore, the bound of \Cref{lemma_eq:22} is analyzed as follows: 
\begin{align}
    &\quad \conE \norm{\sum_{i \in \mathcal{C}^t} \frac{S}{C} \alpha_i + \sum_{i \in \mathcal{R}^t} \frac{N-S}{K-C} \alpha_i} \\
    &= \conE \left(\frac{C}{S} \cdot \left(\frac{S}{C}\right)^2 \sum_{i \in \sticky{t}} \norm{\alpha_i} + \frac{K-C}{N-S} \left(\frac{N-S}{K-C}\right)^2 \sum_{i \in \nonsticky{t}} \norm{\alpha_i} + \frac{C(C-1)}{S^2}\left(\frac{S}{C}\right)^2 \sum_{i, j \in \sticky{t}} \innerp{\alpha_{i}}{\alpha_{j}} \right. \nonumber \\
    &\qquad \qquad \left. + \frac{(K-C)(K-C-1)}{(N-S)^2}\left(\frac{N-S}{K-C}\right)^2 \sum_{i, j \in \nonsticky{t}} \innerp{\alpha_{i}}{\alpha_{j}} + 2 \frac{(K-C)C}{(N-S)S}\left(\frac{S}{C}\right) \left(\frac{N-S}{K-C}\right) \sum_{i \in \sticky{t}, j \in \nonsticky{t}} \innerp{\alpha_i}{\alpha_j} \right) \\
    &= \conE \left(\frac{C}{S} \cdot \left(\frac{S}{C}\right)^2 \sum_{i \in \sticky{t}} \norm{\alpha_i} + \frac{K-C}{N-S} \left(\frac{N-S}{K-C}\right)^2 \sum_{i \in \nonsticky{t}} \norm{\alpha_i} + \frac{C(C-1)}{S^2}\left(\frac{S}{C}\right)^2 \norm{\sum_{i \in \sticky{t}} \alpha_i} \right. \nonumber \\
    &\qquad \qquad \left. + \frac{(K-C)(K-C-1)}{(N-S)^2}\left(\frac{N-S}{K-C}\right)^2 \norm{\sum_{i \in \nonsticky{t}} \alpha_i} + 2 \frac{(K-C)C}{(N-S)S}\left(\frac{S}{C}\right) \left(\frac{N-S}{K-C}\right) \sum_{i \in \sticky{t}, j \in \nonsticky{t}} \innerp{\alpha_i}{\alpha_j} \right) \\
    &\leq \conE \left(\frac{S}{C} \sum_{i \in \sticky{t}} \norm{\alpha_i} + \frac{N-S}{K-C} \sum_{i \in \nonsticky{t}} \norm{\alpha_i} + \norm{\sum_{i=1}^N \alpha_i} \right) \\
    &\leq \conE \left(\frac{S}{C} \sum_{i \in \sticky{t}} p_i^2 E^2 L_c^2 + \frac{N-S}{K-C} \sum_{i \in \nonsticky{t}} p_i^2 E^2 L_c^2 + \norm{\sum_{i=1}^N \alpha_i} \right) \label{eq:32}
\end{align}
where the first equation is due to the independent client sampling with replacement in both groups, and the last inequality follows \Cref{ass:L-smooth}. Therefore, with the result from \Cref{eq:32}, we can obtain the desired result based on \Cref{eq:00014}.

\section{Aggregation for Batch Normalization layers in \mytitle}
\label{sec:bn}
A Batch Normalization (BN) layer contains five parameters: trainable \texttt{weight}, \texttt{bias} layers, and non-trainable summary statistics \texttt{running\_mean}, \texttt{running\_var}, and \texttt{num\_batches\_tracked}. While \mytitle updates trainable parameters (\texttt{weight} and \texttt{bias}) as all model parameters (\Cref{algo:glueFL}), non-trainable parameters (\texttt{running\_mean}, \texttt{running\_var} and \texttt{num\_batches\_tracked}) need to be treated differently. We perform the aggregation of these non-trainable parameters $\mathbf{v}$ as follows:
\begin{gather}
    \Delta_i^{t} \gets \mathbf{v}^{t,E}_i - \mathbf{v}^{t,0}_i \\
    \mathbf{v}^{t+1} \gets \mathbf{v}^{t} + \frac{1}{K} \sum_{i \in \mathcal{K}} \Delta_i^{t}
\end{gather}
where $\Delta_i^{t}$ represents the local change of $\mathbf{v}^{t}$ on client $i$ in round $t$. Note that we do not perform re-weighting on $\Delta_i^{t}$ as this produces the best empirical results. This aggregation rule is consistent with the FedScale implementation~\cite{fedscale-icml22}. 



 


\end{proof}
\end{lemma}

\end{document}